\documentclass[preprint, times]{elsarticle}
\pdfoutput=1
\usepackage{microtype}
\usepackage{fancyvrb}
\usepackage{xcolor}
\usepackage[ruled]{algorithm2e}
\usepackage{amsmath}
\usepackage{booktabs}
\usepackage{makecell}

\usepackage{natbib}
\usepackage{hyperref}
\usepackage{url}
\usepackage{caption}
\usepackage{subcaption}
\biboptions{sort&compress}

\makeatletter
\def\PY@reset{\let\PY@it=\relax \let\PY@bf=\relax%
    \let\PY@ul=\relax \let\PY@tc=\relax%
    \let\PY@bc=\relax \let\PY@ff=\relax}
\def\PY@tok#1{\csname PY@tok@#1\endcsname}
\def\PY@toks#1+{\ifx\relax#1\empty\else%
    \PY@tok{#1}\expandafter\PY@toks\fi}
\def\PY@do#1{\PY@bc{\PY@tc{\PY@ul{%
    \PY@it{\PY@bf{\PY@ff{#1}}}}}}}
\def\PY#1#2{\PY@reset\PY@toks#1+\relax+\PY@do{#2}}

\expandafter\def\csname PY@tok@w\endcsname{\def\PY@tc##1{\textcolor[rgb]{0.73,0.73,0.73}{##1}}}
\expandafter\def\csname PY@tok@c\endcsname{\let\PY@it=\textit\def\PY@tc##1{\textcolor[rgb]{0.25,0.50,0.50}{##1}}}
\expandafter\def\csname PY@tok@cp\endcsname{\def\PY@tc##1{\textcolor[rgb]{0.74,0.48,0.00}{##1}}}
\expandafter\def\csname PY@tok@k\endcsname{\let\PY@bf=\textbf\def\PY@tc##1{\textcolor[rgb]{0.00,0.50,0.00}{##1}}}
\expandafter\def\csname PY@tok@kp\endcsname{\def\PY@tc##1{\textcolor[rgb]{0.00,0.50,0.00}{##1}}}
\expandafter\def\csname PY@tok@kt\endcsname{\def\PY@tc##1{\textcolor[rgb]{0.69,0.00,0.25}{##1}}}
\expandafter\def\csname PY@tok@o\endcsname{\def\PY@tc##1{\textcolor[rgb]{0.40,0.40,0.40}{##1}}}
\expandafter\def\csname PY@tok@ow\endcsname{\let\PY@bf=\textbf\def\PY@tc##1{\textcolor[rgb]{0.67,0.13,1.00}{##1}}}
\expandafter\def\csname PY@tok@nb\endcsname{\def\PY@tc##1{\textcolor[rgb]{0.00,0.50,0.00}{##1}}}
\expandafter\def\csname PY@tok@nf\endcsname{\def\PY@tc##1{\textcolor[rgb]{0.00,0.00,1.00}{##1}}}
\expandafter\def\csname PY@tok@nc\endcsname{\let\PY@bf=\textbf\def\PY@tc##1{\textcolor[rgb]{0.00,0.00,1.00}{##1}}}
\expandafter\def\csname PY@tok@nn\endcsname{\let\PY@bf=\textbf\def\PY@tc##1{\textcolor[rgb]{0.00,0.00,1.00}{##1}}}
\expandafter\def\csname PY@tok@ne\endcsname{\let\PY@bf=\textbf\def\PY@tc##1{\textcolor[rgb]{0.82,0.25,0.23}{##1}}}
\expandafter\def\csname PY@tok@nv\endcsname{\def\PY@tc##1{\textcolor[rgb]{0.10,0.09,0.49}{##1}}}
\expandafter\def\csname PY@tok@no\endcsname{\def\PY@tc##1{\textcolor[rgb]{0.53,0.00,0.00}{##1}}}
\expandafter\def\csname PY@tok@nl\endcsname{\def\PY@tc##1{\textcolor[rgb]{0.63,0.63,0.00}{##1}}}
\expandafter\def\csname PY@tok@ni\endcsname{\let\PY@bf=\textbf\def\PY@tc##1{\textcolor[rgb]{0.60,0.60,0.60}{##1}}}
\expandafter\def\csname PY@tok@na\endcsname{\def\PY@tc##1{\textcolor[rgb]{0.49,0.56,0.16}{##1}}}
\expandafter\def\csname PY@tok@nt\endcsname{\let\PY@bf=\textbf\def\PY@tc##1{\textcolor[rgb]{0.00,0.50,0.00}{##1}}}
\expandafter\def\csname PY@tok@nd\endcsname{\def\PY@tc##1{\textcolor[rgb]{0.67,0.13,1.00}{##1}}}
\expandafter\def\csname PY@tok@s\endcsname{\def\PY@tc##1{\textcolor[rgb]{0.73,0.13,0.13}{##1}}}
\expandafter\def\csname PY@tok@sd\endcsname{\let\PY@it=\textit\def\PY@tc##1{\textcolor[rgb]{0.73,0.13,0.13}{##1}}}
\expandafter\def\csname PY@tok@si\endcsname{\let\PY@bf=\textbf\def\PY@tc##1{\textcolor[rgb]{0.73,0.40,0.53}{##1}}}
\expandafter\def\csname PY@tok@se\endcsname{\let\PY@bf=\textbf\def\PY@tc##1{\textcolor[rgb]{0.73,0.40,0.13}{##1}}}
\expandafter\def\csname PY@tok@sr\endcsname{\def\PY@tc##1{\textcolor[rgb]{0.73,0.40,0.53}{##1}}}
\expandafter\def\csname PY@tok@ss\endcsname{\def\PY@tc##1{\textcolor[rgb]{0.10,0.09,0.49}{##1}}}
\expandafter\def\csname PY@tok@sx\endcsname{\def\PY@tc##1{\textcolor[rgb]{0.00,0.50,0.00}{##1}}}
\expandafter\def\csname PY@tok@m\endcsname{\def\PY@tc##1{\textcolor[rgb]{0.40,0.40,0.40}{##1}}}
\expandafter\def\csname PY@tok@gh\endcsname{\let\PY@bf=\textbf\def\PY@tc##1{\textcolor[rgb]{0.00,0.00,0.50}{##1}}}
\expandafter\def\csname PY@tok@gu\endcsname{\let\PY@bf=\textbf\def\PY@tc##1{\textcolor[rgb]{0.50,0.00,0.50}{##1}}}
\expandafter\def\csname PY@tok@gd\endcsname{\def\PY@tc##1{\textcolor[rgb]{0.63,0.00,0.00}{##1}}}
\expandafter\def\csname PY@tok@gi\endcsname{\def\PY@tc##1{\textcolor[rgb]{0.00,0.63,0.00}{##1}}}
\expandafter\def\csname PY@tok@gr\endcsname{\def\PY@tc##1{\textcolor[rgb]{1.00,0.00,0.00}{##1}}}
\expandafter\def\csname PY@tok@ge\endcsname{\let\PY@it=\textit}
\expandafter\def\csname PY@tok@gs\endcsname{\let\PY@bf=\textbf}
\expandafter\def\csname PY@tok@gp\endcsname{\let\PY@bf=\textbf\def\PY@tc##1{\textcolor[rgb]{0.00,0.00,0.50}{##1}}}
\expandafter\def\csname PY@tok@go\endcsname{\def\PY@tc##1{\textcolor[rgb]{0.53,0.53,0.53}{##1}}}
\expandafter\def\csname PY@tok@gt\endcsname{\def\PY@tc##1{\textcolor[rgb]{0.00,0.27,0.87}{##1}}}
\expandafter\def\csname PY@tok@err\endcsname{\def\PY@bc##1{\setlength{\fboxsep}{0pt}\fcolorbox[rgb]{1.00,0.00,0.00}{1,1,1}{\strut ##1}}}
\expandafter\def\csname PY@tok@kc\endcsname{\let\PY@bf=\textbf\def\PY@tc##1{\textcolor[rgb]{0.00,0.50,0.00}{##1}}}
\expandafter\def\csname PY@tok@kd\endcsname{\let\PY@bf=\textbf\def\PY@tc##1{\textcolor[rgb]{0.00,0.50,0.00}{##1}}}
\expandafter\def\csname PY@tok@kn\endcsname{\let\PY@bf=\textbf\def\PY@tc##1{\textcolor[rgb]{0.00,0.50,0.00}{##1}}}
\expandafter\def\csname PY@tok@kr\endcsname{\let\PY@bf=\textbf\def\PY@tc##1{\textcolor[rgb]{0.00,0.50,0.00}{##1}}}
\expandafter\def\csname PY@tok@bp\endcsname{\def\PY@tc##1{\textcolor[rgb]{0.00,0.50,0.00}{##1}}}
\expandafter\def\csname PY@tok@fm\endcsname{\def\PY@tc##1{\textcolor[rgb]{0.00,0.00,1.00}{##1}}}
\expandafter\def\csname PY@tok@vc\endcsname{\def\PY@tc##1{\textcolor[rgb]{0.10,0.09,0.49}{##1}}}
\expandafter\def\csname PY@tok@vg\endcsname{\def\PY@tc##1{\textcolor[rgb]{0.10,0.09,0.49}{##1}}}
\expandafter\def\csname PY@tok@vi\endcsname{\def\PY@tc##1{\textcolor[rgb]{0.10,0.09,0.49}{##1}}}
\expandafter\def\csname PY@tok@vm\endcsname{\def\PY@tc##1{\textcolor[rgb]{0.10,0.09,0.49}{##1}}}
\expandafter\def\csname PY@tok@sa\endcsname{\def\PY@tc##1{\textcolor[rgb]{0.73,0.13,0.13}{##1}}}
\expandafter\def\csname PY@tok@sb\endcsname{\def\PY@tc##1{\textcolor[rgb]{0.73,0.13,0.13}{##1}}}
\expandafter\def\csname PY@tok@sc\endcsname{\def\PY@tc##1{\textcolor[rgb]{0.73,0.13,0.13}{##1}}}
\expandafter\def\csname PY@tok@dl\endcsname{\def\PY@tc##1{\textcolor[rgb]{0.73,0.13,0.13}{##1}}}
\expandafter\def\csname PY@tok@s2\endcsname{\def\PY@tc##1{\textcolor[rgb]{0.73,0.13,0.13}{##1}}}
\expandafter\def\csname PY@tok@sh\endcsname{\def\PY@tc##1{\textcolor[rgb]{0.73,0.13,0.13}{##1}}}
\expandafter\def\csname PY@tok@s1\endcsname{\def\PY@tc##1{\textcolor[rgb]{0.73,0.13,0.13}{##1}}}
\expandafter\def\csname PY@tok@mb\endcsname{\def\PY@tc##1{\textcolor[rgb]{0.40,0.40,0.40}{##1}}}
\expandafter\def\csname PY@tok@mf\endcsname{\def\PY@tc##1{\textcolor[rgb]{0.40,0.40,0.40}{##1}}}
\expandafter\def\csname PY@tok@mh\endcsname{\def\PY@tc##1{\textcolor[rgb]{0.40,0.40,0.40}{##1}}}
\expandafter\def\csname PY@tok@mi\endcsname{\def\PY@tc##1{\textcolor[rgb]{0.40,0.40,0.40}{##1}}}
\expandafter\def\csname PY@tok@il\endcsname{\def\PY@tc##1{\textcolor[rgb]{0.40,0.40,0.40}{##1}}}
\expandafter\def\csname PY@tok@mo\endcsname{\def\PY@tc##1{\textcolor[rgb]{0.40,0.40,0.40}{##1}}}
\expandafter\def\csname PY@tok@ch\endcsname{\let\PY@it=\textit\def\PY@tc##1{\textcolor[rgb]{0.25,0.50,0.50}{##1}}}
\expandafter\def\csname PY@tok@cm\endcsname{\let\PY@it=\textit\def\PY@tc##1{\textcolor[rgb]{0.25,0.50,0.50}{##1}}}
\expandafter\def\csname PY@tok@cpf\endcsname{\let\PY@it=\textit\def\PY@tc##1{\textcolor[rgb]{0.25,0.50,0.50}{##1}}}
\expandafter\def\csname PY@tok@c1\endcsname{\let\PY@it=\textit\def\PY@tc##1{\textcolor[rgb]{0.25,0.50,0.50}{##1}}}
\expandafter\def\csname PY@tok@cs\endcsname{\let\PY@it=\textit\def\PY@tc##1{\textcolor[rgb]{0.25,0.50,0.50}{##1}}}

\def\PYZus{\char`\_}

\def\PYZsh{\char`\#}

\def\PYZhy{\char`\-}

\makeatother

\title{
    Maximizing Model Generalization for
    Machine Condition Monitoring
    with Self-Supervised Learning
    and Federated Learning
}
\author[1]{Matthew Russell}
\author[1,2]{Peng Wang\corref{cor1}}
\affiliation[1]{
    organization={
            Department of Electrical and Computer Engineering,
            University of Kentucky},
    city={Lexington},
    state={KY},
    country={USA}
}
\affiliation[2]{
    organization={
            Department of Mechanical and Aerospace Engineering,
            University of Kentucky},
    city={Lexington},
    state={KY},
    country={USA}
}
\cortext[cor1]{Corresponding author}
\begin{document}
\begin{abstract}
    Deep Learning (DL) can diagnose faults
    and assess machine health
    from raw condition monitoring data
    without manually designed statistical features.
    However, practical manufacturing applications
    require robust and repeatable solutions that can be trusted in dynamic environments.
    Machine data is often unlabeled
    and from very few health conditions
    (e.g., only normal operating data).
    Furthermore, models often encounter
    shifts in domain as process parameters change
    and new categories of faults emerge.
    Traditional supervised learning may struggle to learn
    compact, discriminative representations that generalize
    to these unseen target domains
    since it depends on having plentiful classes
    to partition the feature space with decision boundaries.
    Transfer Learning (TL) with domain adaptation attempts
    to adapt these models to unlabeled target domains
    but assumes similar underlying structure
    that may not be present if new faults emerge.
    This study proposes focusing on maximizing
    the feature generality on the source domain
    and applying TL via weight transfer
    to copy the model to the target domain.
    Specifically, Self-Supervised Learning (SSL)
    with Barlow Twins may produce more discriminative features
    for monitoring health condition
    than supervised learning by focusing on
    semantic properties of the data.
    Furthermore, Federated Learning (FL)
    for distributed training may also improve generalization
    by efficiently expanding the effective size and diversity
    of training data by sharing information
    across multiple client machines.
    Results show that Barlow Twins outperforms
    supervised learning in an unlabeled target domain
    with emerging motor faults
    when the source training data contains very few distinct categories.
    Incorporating FL may also provide a slight advantage
    by diffusing knowledge of health conditions between machines.
    Future work should continue investigating SSL and FL performance
    in these realistic manufacturing scenarios.
\end{abstract}

\maketitle

\section{Introduction}
Smart factories need to detect and diagnose machine faults
to prevent costly downtime and repairs.
To this end,
machine learning can build classification and regression models
for condition monitoring and fault diagnosis
using statistical patterns
discovered in large data sets.
Deep Learning (DL) has shifted the paradigm
away from manually-designed features
(e.g., mean, variance, kurtosis, peak values, etc.)
by introducing efficient algorithms
for training neural networks with many layers
to extract features automatically from raw data
(e.g., vibration signals)~\cite{krizhevsky2012_nips, lecun2015_nature}.
However, practical manufacturing applications
require robust and repeatable solutions
that can be trusted in dynamic environments.
Since DL models derive their behavior from empirical training data,
predictions can be difficult to verify and validate,
and large quantities of clean examples are not available
covering all possible operating conditions and process parameters.
Building such exhaustive training sets is prohibitively time-consuming and cost-intensive---not
to mention the privacy concerns if data comes from many different sources.
Widespread use of DL for condition monitoring
hinges on finding effective alternatives that promote trust and repeatability.

Interest in pursuing trustworthy DL stems from early DL
work in manufacturing
that
established its superiority
over traditional approaches
like Support Vector Machine (SVM)
for analyzing condition monitoring data sets~\cite{wen2017_tie}.
Despite excellent results
on controlled laboratory data sets,
many practical considerations hinder widespread
adoption of DL within manufacturing.
Contrary to image domains that have millions of images
from hundreds or thousands of categories~\cite{krizhevsky2012_nips},
fault diagnosis problems often lack the volume and diversity
of data required to learn robust feature extraction networks
that generalize beyond a single data set,
operating condition, or machine~\citep{li2020_jim}.
Healthy operating conditions dominate
real-world industrial data sets
with very few---if any---examples of faults~\citep{sun2022_tie}.
Any limited examples of faults will be overwhelmingly unlabeled.
Furthermore,
factory
environments are dynamic;
new types of faults
can occur without warning
and be confidently misclassified
by an outdated model~\cite{yu2019_tii, li2021_mech, fu2022_mssp}.
Thus, increased operational trust starts with improving generalization to ensure models behave more predictably
when confronted with uncertain process dynamics
and incomplete observational knowledge.

Transfer Learning (TL) can alleviate some issues with generalization.
TL seeks to repurpose and reuse a model
when faced with changing data or tasks
(e.g., new faults or process parameters)~\cite{pan2010_tkde}.
These changes affect the statistical properties of the data,
shifting it out of the model's valid input domain~\cite{kouw2018_arxiv}.
TL for domain adaptation transfers a model
from a labeled source domain
to an unlabeled target domain.
However, emerging faults in the target domain
may hinder the ability to transfer the source domain model.
Additionally,
the target domain itself could be unknown
or represent a future operating state
with no data---even unlabeled data---available
at training time.
In this case, TL approaches must
learn the most generalizable
representation possible from the available data.
The model can then be transferred to the target domain
and used as-is or fine tuned as target domain data
becomes available~\cite{pan2010_tkde}.
This technique can bootstrap models for the target domain
without assuming an isomorphic relationship
to the source domain conditions.

Bootstrapping source models with supervised learning (i.e., labeled data)
may be ineffective in practical condition monitoring
since few training conditions are available,
and labels are often missing.
Self-Supervised Learning (SSL) may be more appropriate.
SSL techniques
create compact clusters of features
with similar semantic
characteristics~\citep{dosovitskiy2014_nips}.
Random augmentations
(e.g., random scale, time shift, etc.)
implicitly specify what variation
the model should expect within a category of signals.
For example, if both a randomly flipped signal and the original
should map to the same feature,
the model learns to ignore flipping.
Requiring no labels,
SSL facilitates learning data-centric representations
from raw, unannotated factory data.

While SSL may better bootstrap condition monitoring models,
generalization can be improved further by sharing information
among a fleet of machines.
Bandwidth constraints may prevent the fleet
from continually aggregating data in the cloud,
but Federated Learning (FL)
can utilize the distributed data efficiently
to develop a globally-informed model~\cite{mcmahan2017_aistats}.
Each client machine trains on locally observed data
and periodically transmits its model---not the raw data---to a server
which combines the updates into a single model.
This global model is then distributed to the clients,
diffusing information among them.
Thus, FL can expand the effective size and diversity data sets
by integrating information from multiple clients
without inundating communication networks.

Condition monitoring literature lacks
a cohesive introduction to SSL and FL
for maximizing model generalization.
This study outlines how SSL and FL
can improve the generalization---and therefore trustworthiness---of
DL models on the factory floor
via two complementary strategies:
SSL extracts informative representations without needing labeled data,
and FL expands the effective size and diversity of the data set.
Pursuing generalizable models through SSL and FL
allows manufacturers to adopt a knowledge-informed approach
and securely share information via FL among clients grouped by expert knowledge
while simultaneously maximizing the utilization of massively unlabeled data via SSL.
The contributions of this study can be summarized as follows:
\begin{enumerate}
    \itemsep0em
    \item an overview of SSL
          and related work in manufacturing,
    \item an overview of FL
          and related work in manufacturing,
    \item a theoretically motivated framework for combining
          SSL and FL to improve model generalization, and
    \item a case study assessing SSL and FL
          under emerging faults and changing process parameters
          using a motor fault data set.
\end{enumerate}
The rest of this paper is organized as follows:
Section~\ref{sec:70-related-work} outlines the theoretical background
and related work,
Section~\ref{sec:70-proposed} descrivbes the proposed SSL and FL
methods for condition monitoring,
Section~\ref{sec:70-experiments} introduces a
motor health condition case study,
Section~\ref{sec:70-results} presents and discusses the results,
and Section~\ref{sec:70-conclusion} provides concluding remarks.

\section{Theoretical Background and Related Work\label{sec:70-related-work}}
This work builds on Transfer Learning, Self-Supervised Learning, and Federated Learning.

\subsection{Supervised Learning and Transfer Learning}
Many factors can limit the applicability and robustness
of machine learning models.
In manufacturing,
changing processing parameters,
operating environments,
and health conditions
can negatively impact performance
by shifting the input data distribution
outside the expected domain.
Transfer Learning (TL) seeks to adapt or reuse
models trained in a source domain
to a related target domain~\cite{pan2010_tkde},
circumventing the need for
large volumes of labeled data for the target task.

\subsubsection{Supervised Learning}

A typical fault diagnosis model can be split into
a feature extraction backbone $G_\theta$
parameterized by weights $\theta$
and classification head $F_\phi$
with weights $\phi$
that predicts the probabilities
of $K$ classes (e.g., faults)
from the extracted features.
With labeled data,
the model parameters can be optimized
with stochastic gradient descent and backpropagation
using the cross-entropy loss (i.e., cost) function:
\begin{equation}
    \label{eq:70-cross-entropy-loss}
    \mathcal L_{\text{CE}}(X, Y)
    = - \frac{1}{n} Y^\top \log F_\phi(G_\theta(X))
\end{equation}
where $X = [\mathbf x_1 \mathbf x_2 \dots \mathbf x_n]$
is a batch of $n$ input examples,
$Y = [\mathbf y_1 \mathbf y_2 \dots \mathbf y_n]$
is corresponding binary label vectors
$\mathbf y \in \{0, 1\}^K$
with $1$ in the index corresponding to the true label
and zeros elsewhere, and
$F_\phi(G_\theta(X)) = [\hat{\mathbf y}_1 \hat{\mathbf y}_2 \dots \hat{\mathbf y}_n]$
is the set of predicted class probabilities for the batch.
Optimizing the weights to maximize classification accuracy
teaches the model to draw ``decision boundaries''
that separate the features from different categories.
However, changes in process parameters or operating environment
shift the distribution of input data and features from $G_\theta$.
These new features no longer align with the decision boundaries learned
by the classifier $F_\phi$, producing undefined or inconsistent behavior.
This damages the generalization of supervised classifiers.

\subsubsection{Transfer Learning via Domain Adaptation}
Transfer Learning (TL) is one solution to the domain shift problem.
For domain adaptation,
unlabeled data from a known target domain can
regularize the supervised training process
so $G_\theta$ produces stable, matching distributions
of source and target domain features for the classifier $F_\phi$.
An updated loss function that includes unlabeled target domain data
is used during training:
\begin{equation}
    \label{eq:70-tl-domain-adaptation}
    \mathcal L_{\text{DA}}(X_s, Y_s, X_t)
    = \mathcal L_{\text{CE}}(X_s, Y_s)
    + \lambda D(G_\theta(X_s), G_\theta(X_t))
\end{equation}
where $X_s$ is the batch of source domain inputs,
$Y_s$ is the batch of source domain labels,
$X_t$ is the batch of unlabeled target domain inputs,
and $D(\cdot, \cdot)$ is a function
measuring the distribution discrepancy between
source domain features $G_\theta(X_s)$ and
target domain features $G_\theta(X_t)$~\cite{wang2020_cirp}.
The $\lambda$ factor controls the strength of feature regularization.
Since the feature extractor $G_\theta$ produces a consistent distribution
of features from both the source and target domains,
the fault classifer $F_\phi$
is more likely to generate accurate predictions
for the target domain.

A popular implementation of $D(\cdot, \cdot)$ in manufacturing
is Maximum Mean Discrepancy (MMD).
Using MMD to ensure similarity between source and target features,
\cite{lu2017_tie} demonstrated TL of bearing and gearbox vibration data
across different loads and shaft speeds.
With flexible kernel implementations,
MMD can be combined with a polynomial or Cauchy kernel
as shown on laboratory fault data sets~\cite{yang2019_tie, cao2022_jms}.
Applying MMD at multiple levels
in a deep feature extractor
can also provide performance gains
for lab-to-real transfer for locomotive bearing fault diagnosis
and classification and localization of bearing faults~\cite{yang2019_mssp, su2022_jms}.

Rather than relying on an explicit metric,
another widely used approach
is Domain Adversarial Neural Network (DANN)
which replaces the $D(\cdot, \cdot)$
loss term with another neural network $D_\psi$ that learns
to discriminate source and target features~\cite{ganin2016_jmlr}.
By training the feature extractor $G_\theta$
to confuse the domain discriminator $D_\psi$,
the feature extractor learns to generate matching
features for source and target domain data.
DANN can facilitate TL across different bearing data sets
with a 1D CNN feature extractor~\cite{li2019_tii}.
Interestingly, combining both MMD and DANN may be beneficial
and has also been demonstrated for TL across data sets~\cite{guo2018_tie}.

\subsubsection{Transfer Learning via Weight Transfer}
Domain adaptation may encounter difficulties when new faults emerge.
If the target domain contains emerging faults,
encouraging source and target features to match may be detrimental.
Furthermore, the classifier itself must be
reconfigured to detect the additional fault(s).
Thus, instead of domain adaptation,
TL under emerging faults
shifts to maximizing the generalization
of feature representations learned on the source domain.
If the representation is general enough,
network weights can be transferred
to the target domain to separate emerging faults
and previously known faults.
That is, given labeled source and/or unlabeled target domain data,
TL via weight transfer seeks to pretrain a representation
that remains discriminative for future emerging faults.
In image processing, weight transfer allows
applications to reuse low-level, general features learned by
networks trained on massive image data sets~\cite{yosinski2014_nips}.
The size and diversity of the training data
enables these pretrained networks to produce
highly discriminative features for emerging categories of images.
Starting from these pretrained weights can produce useful feature representations
for solving problems in domains like medical imaging
where data is too scarce to train
reliable image classifiers from scratch~\cite{zhuang2020_ieee}.

Manufacturing researchers have leveraged these pretrained
image networks creatively
by transforming condition monitoring data sets into images.
While the high-level tasks differ,
pretrained networks extract
useful low-level information about lines and shapes
in the images~\cite{krizhevsky2012_nips}.
If vibration data is transformed into 2D images
via the Continuous Wavelet Transform (CWT),
these pretrained image networks
can provide out-of-the-box features
for training fault classifiers
when labeled manufacturing data is limited~\cite{shao2018_tii, wang2020_ress}.
They can even accelerate domain adaptation
by providing the initial feature representation
before applying a technique like MMD~\cite{wang2020_cirp}.
Outside of pretrained image networks,
\cite{he2021_tie} demonstrated that TL via weight transfer
can improve predictions of a target aircraft engine's degradation
by training a degradation model on source engines,
transfering the weights to the target engine,
and then fine-tuning on the target's first few degradation steps.
However, in many cases TL via weight transfer
remains difficult for manufacturing
because of the lack of labeled data required to pretrain
highly general feature extractors.

\subsection{Self-Supervised Learning}

Self-Supervised Learning (SSL)
uses unlabeled data to train feature extraction networks
that can be transferred to downstream tasks.
Broadly speaking, SSL lets the data ``supervise itself''
through pretext tasks or invariance-based methods
to learn a useful encoding of the input examples.
SSL could be transformational in manufacturing
where labeled data is scarce
and unlabeled data is plentiful.

\subsubsection{Pretext Task SSL}
Pretext task SSL trains models
on a related problem using autogenerated labels.
Examples of pretext tasks include
predicting image rotations~\cite{gidaris2018_iclr},
the relative position of patches within an image~\cite{doesrch2015_iccv},
or the next word in a natural language sequence
(e.g., GPT-\textit{n} models from OpenAI)~\cite{radford2018_openai}.
Manufacturing and health monitoring research has explored
various adaptations of this approach.
Some studies rebrand
traditional unsupervised techniques
as ``self-supervised.''
For example,
an embedding learned from only normal data via
Kernel Principal Component Analysis (PCA)
helped detect faults
in an industrial metal etching process
and was described as self-supervised~\cite{wang2018_jim}.
Similarly, \cite{zhang2021_sensors}
trained a deep autoencoder
as a ``self-supervised'' auxiliary task
for bearing fault classification,
while \cite{shul2023_mssp} adopted a similar approach
for anomaly detection in washing machines.
Work by \cite{kim2023_jms} predicted the orientation of randomly rotated
laser powder bed fusion process images from additive manufacturing
and characterized this as a pretext task.
However, since the downstream task was also orientation prediction,
this resembles pretraining with data augmentation
rather than a distinct pretext goal.
True pretext task SSL
for downstream fault diagnosis
mines features from unlabeled data
via distinct pretraining tasks
that do not depend on fault information.
For example,
a model could learn useful features by
predicting statistical properties
of unlabeled input signals
(e.g., mean, variance, skew, and kurtosis)~\cite{zhang2022_tie}.
Both \cite{wang2022_kbs} and \cite{nie2023_sensors}
randomly distorted input signals and
trained a model to identify the applied distortion.
All three approaches produced features useful for bearing fault diagnosis.
Thus, without requiring manual labels,
pretext task SSL can bootstrap models
for future health monitoring tasks.

\subsubsection{Invariance-Based SSL}
\begin{figure}[t]
    \centering
    \includegraphics[width=3.25in]{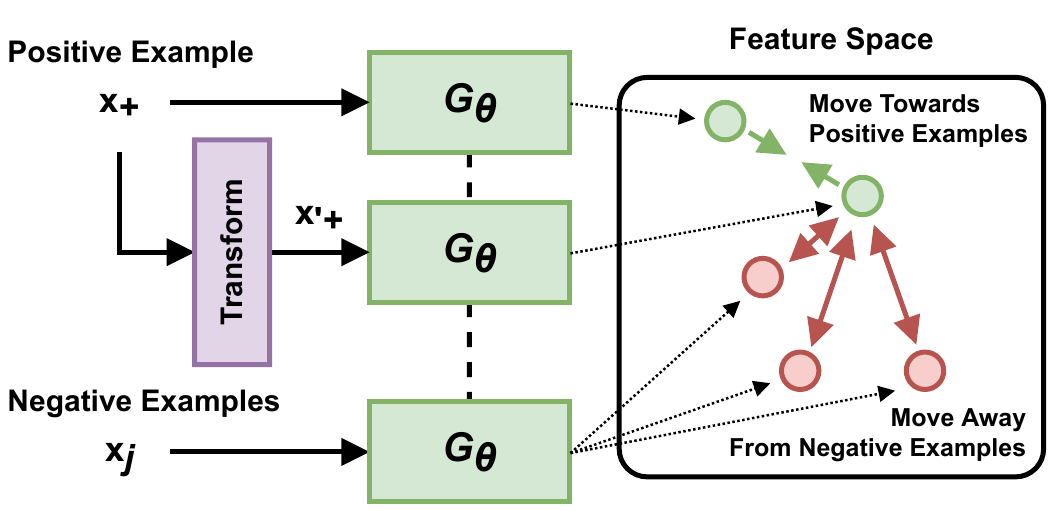}
    \caption{SSL techniques seek to move augmented features
        towards members of the same pseudoclass
        while increasing separation
        from other pseudoclasses.\label{fig:70-ssl}}
\end{figure}

Instead of using pretext tasks,
invariance-based SSL applies random transformations
to a ``seed'' example from the data set,
creating family of examples belonging to the same ``pseudoclass.''
The feature extraction network is then trained to homogenize
features from all augmented examples in the pseudoclass~\cite{dosovitskiy2014_nips}.
A contrastive loss function encourages each pseudoclass
to be both compact and well-separated from others~\cite{hadsell2006_cvpr}.
Through this process, the network learns to ignore
the randomized attributes and focus on semantically meaningful
ways to cluster the inputs data (see Fig.~\ref{fig:70-ssl}).

Contrastive approaches to Invariance-based SSL
depend on having plentiful ``negative'' examples
of other pseudoclasses to ensure good clustering.
For example, consider the InfoNCE loss function,
where $\mathbf x'_+$ is an augmented version (i.e., same pseudoclass)
of a positive reference example $\mathbf x_+$~\cite{oord2018_arxiv,he2020_cvpr}:
\begin{equation}
    \label{eq:70-infonce-loss}
    \mathcal L_{\text{InfoNCE}} =
    - \log \frac{
        s(G_\theta(\mathbf x'_+), G_\theta(\mathbf x_+))}
    {\sum_{j=1}^n s(G_\theta(\mathbf x'_+), G_\theta(\mathbf x_j))}
\end{equation}
where $n$ is the size of the batch that includes a positive example $\mathbf x_+$
and $n - 1$ negative examples (i.e., other pseudoclasses),
and $s(\cdot, \cdot)$ is a similarity metric.
Increasing the number of negative examples
increases the lower bound on the mutual information (similarity)
between the features of the positive sample $G_\theta(\mathbf x_+)$
and those of its augmentation $G_\theta(\mathbf x'_+)$~\cite{oord2018_arxiv}.
This will encourage compact feature clusters.
However, efficiently training with enough
negative examples can be nontrivial
since the batch size is limited~\cite{hermans2017_arxiv}.
Momentum Contrast (MoCo)~\cite{he2020_cvpr} increased the number of negative examples
by accumulating features across multiple batches.
The encoder trained with contrastive loss
to separate the current batch
from this larger group of negative example features.
A ``momentum encoder'' embeded the previous examples
into the latent space was updated through a running average
to ensure the representations of negative examples
from multiple previous batches remained stable.

\begin{figure}[t]
    \centering
    \includegraphics[width=3.25in]{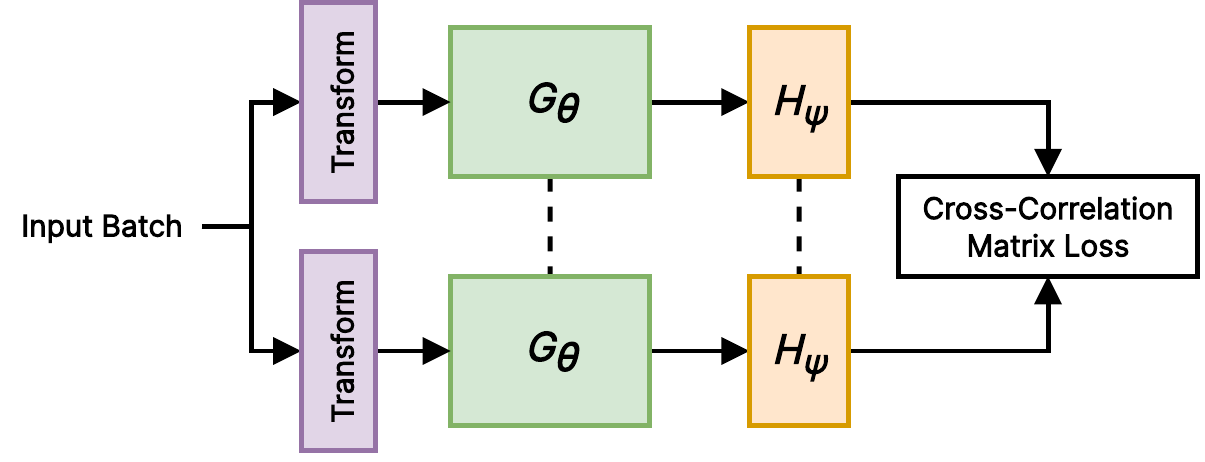}
    \caption{Barlow Twins encourages feature projections to be correlated within each pseudoclass
        and independent from each other to reduce redundancy in the representation.\label{fig:barlow-twins}}
\end{figure}

MoCo prompted many conceptually related developments.
A Simple Framework for Contrastive Learning
of Visual Representations (SimCLR)~\cite{chen2020_icml}
and Bootstrap Your Own Latent (BYOL)~\cite{grill2020_nips}
both proposed modifications of the MoCo-style architecture
that could perform well with fewer or no negative examples.
SimCLR made the important contribution of a ``projection head''
network that mapped features to a larger-dimension space
before applying contrastive loss,
protecting the features themselves from being too aggressively homogenized.
Work by \cite{chen2021_cvpr} proposed an even more straightforward approach
known as Simple Siamese Representation Learning (SimSiam).
SimSiam learned to consolidate feature projections from two augmentations
while preventing gradients from one of the projections
from updating the encoder.
This effectively held one projection stationary
while moving the other towards this anchor.
This proved effective even without large batches,
plentiful negative examples, or momentum networks.
Bypassing issues with contrastive loss altogether,
Barlow Twins used a cross-correlation loss that learned
correlated features among pseudoclass examples
while discouraging redundancy among the feature dimensions
(see Fig.~\ref{fig:barlow-twins})~\cite{zbontar2021_icml}.
Subsequently, Variance-Invariance-Covariance Regularization (VICReg)
introduced a generalization of Barlow Twins
with a slightly more complex loss function~\cite{bardes2022_iclr}.
These methods proved increasingly useful for computer vision problems.

Within manufacturing,
invariance-based SSL from computer vision
can be leveraged
by first converting 1D sensing data into 2D images.
With 2D images of unlabeled vibration data,
SimCLR can find
discriminative fault features for rotating machinery
using image augmentations like rotations, crops, and affine transforms~\cite{wei2022_sensors}.
Utilizing BYOL, \cite{zhang2023_tii}
extracted bearing fault features
after converting vibration data
to images with methods including Short-Time Fourier Transform (STFT)
and Continous Wavelet Transform (CWT).
However, applying image domain techniques to vibration data
might lack a robust, physically meaningful interpretation.
Therefore, an important step for adapting invariance-based SSL
to condition monitoring
is designing appropriate random augmentations for raw time series data
(e.g., vibration and electrical current) that guide training
toward features with rich fault information.

\subsubsection{Designing Time Series Data Augmentations}
\begin{figure}[t]
    \centering
    \includegraphics[width=3.25in]{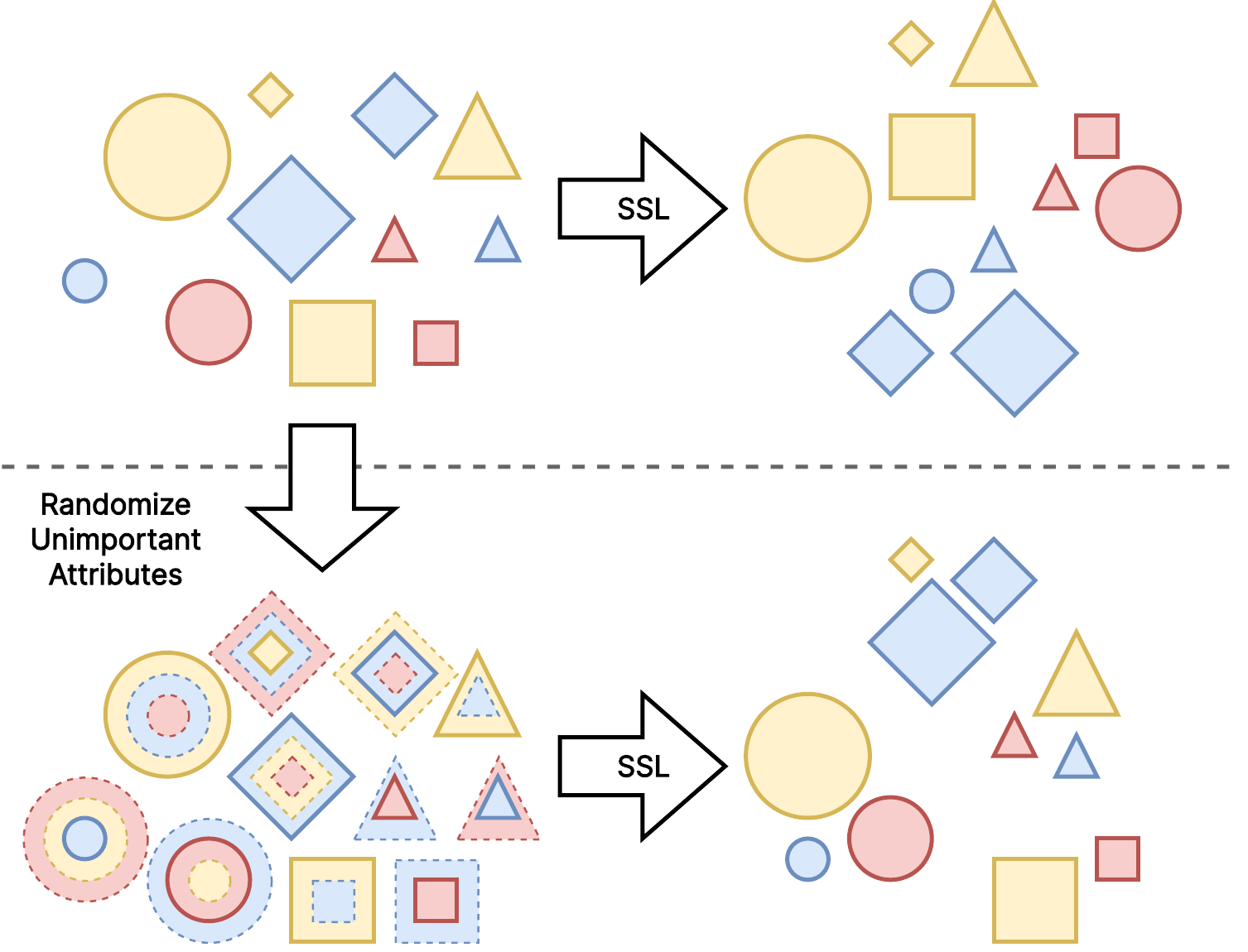}
    \caption{By randomizing semantically meaningless attributes,
        augmentations force SSL to identify pseudoclasses through
        the remaining, semantically meaningful characteristics.\label{fig:augmentation}}
\end{figure}

The random augmentations used by invariance-based SSL
must be carefully selected
to avoid destroying important semantic information.
Input semantics often emerge from complex underlying relationships;
high-level, semantic labels (e.g., bearing inner race fault) cannot be reduced
to a simple feature analysis (e.g., normalized vibration amplitude exceeding 0.6)
nor should this be expected.
The difficulty in uncovering these nonobvious correlations
motivates the use of DL.
Therefore, if an input attribute is semantically meaningful,
extracting and manipulating the attribute tends to be very difficult
(e.g., algorithmically transforming vibrations
from a bearing inner race fault to a healthy vibration signal).
The contrapositive is also true:
if an attribute is not difficult to manipulate,
it will likely not be semantically meaningful (to an extent).
Thus, effective random augmentations need not be complex
to homogenize representations of semantically-related examples
(see Fig.~\ref{fig:augmentation}).
Existing augmentation-based SSL work with images supports this theory
by using simple transforms like
translation, crop, flip, rotation, contrast, blur, and color distortion
for state-of-the-art
results~\cite{chen2021_cvpr, zbontar2021_icml, bardes2022_iclr}.
Each domain is different~\cite{balestriero2023_arxiv},
and designing equivalent augmentations for 1D time series data
unlocks the potential of invariance-based SSL
for raw sensing signals.

Several studies have explored possible time series augmentations.
Since time series examples are related temporally (unlike images),
\cite{hu2023_tii} generated pseudoclasses for invariance-based SSL
from pairs of consecutive instances from the vibration signal
in addition to time and amplitude distortions of single instances.
Gaussian noise, amplitude scaling, stretching, masking, and time shifting
were used with MoCo to pretrain a feature extractor
for detect incipient faults in bearing histories~\cite{ding2022_ress}.
Adopting BYOL, \cite{peng2022_tie}
used truncation (i.e., masking a contiguous region),
lowpass filtering,
Gaussian noise,
geometric scaling,
and downsampling
to learn representations from raw, unlabeled vibration data
for bearing fault diagnosis.
Results indicated that truncation and downsampling were particularly useful.
A similar study utilizing SimSiam was conducted by \cite{wan2023_nnls}
with truncation, lowpass filtering, Gaussian noise, and time reversal.
Using a motor condition data set,
\cite{russell2023_tie} implemented Barlow Twins
on multichannel vibration and current signals
with random time shifting, truncation, scaling, and vertical flipping.
The random time shift was crucial
for extracting good features
for the motor fault diagnosis task.
These studies demonstrate effective data augmentations
when applying invariance-based SSL to 1D signals.

\subsection{Federated Learning}

\begin{figure*}[t]
    \centering
    \includegraphics[width=\textwidth]{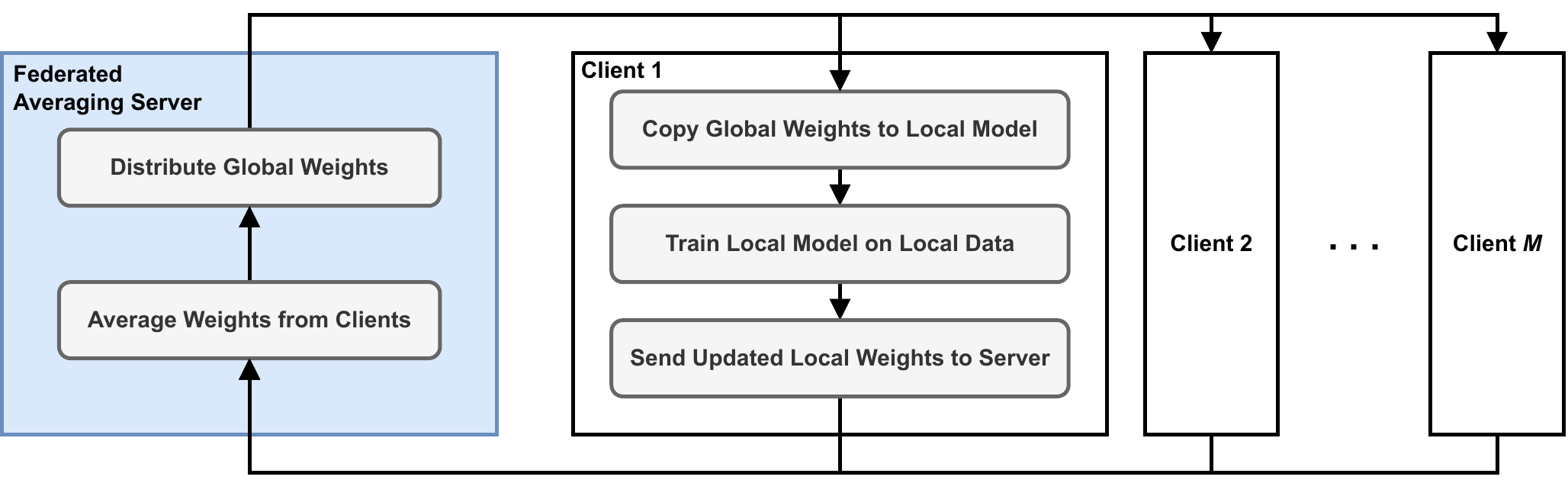}
    \caption{Overview of Federated Learning using
        \texttt{FedAvg}\label{fig:70-fedavg}}
\end{figure*}

Federated Learning (FL) facilitates
distributed training of predictive deep learning models
on private user data via the \texttt{FedAvg} algorithm~\cite{mcmahan2017_aistats}.
To maintain user privacy,
network training is performed on the user's device---only
the updated model weights and parameters
are sent to the cloud.
In the \texttt{FedAvg} algorithm,
the network weights are averaged together to create the global model
without needing to send any client data to the cloud.
This allows clients to retain private control over their data
while still collaborating to train a more generalizable model.
Algorithm~\ref{alg:70-fedavg} outlines
\texttt{FedAvg},
starting with a randomized global model $w^0$
and performing $N$ rounds of federation.
The global model for round $i = 1, 2, \ldots, N$
is distributed to $M$ clients
who update the model using $n$ local minibatches of data.
Each client $j = 1, 2, \ldots, M$ then transmits
the updated model $w_j^i$ for round $i$
back to the server.
Each client also transmits
total amount of training data $a_j$ used by client $j$.
Once all the updates are received for round $i$,
the server computes the global model via weighted average:
\begin{equation}
    \label{eq:70-fedavg}
    w^i = \frac{1}{\sum_{l=1}^M a_l} \sum_{j=1}^M a_j w^i_j
\end{equation}
The weighting coefficients $a_j$ ensure that the global update
is biased towards client models that trained on more data,
which are likely to produce a more stable step
than models trained on only a few examples.
The global model is then redistributed to all the clients
for the next round of FL.

\begin{algorithm}[t]
    \SetKwInOut{Input}{Input}\SetKwInOut{Output}{Output}

    \caption{The \texttt{FedAvg}
        FL algorithm~\cite{mcmahan2017_aistats}}\label{alg:70-fedavg}

    \Input{Number of rounds $N$; number of clients $M$; client steps per round $n$;}
    \Output{Trained global model weights $w^N$}
    $w^0 \gets$ Random initial model weights\;
    \For{$i \gets 1$ \KwTo{} $N$}{
        \For{$j \gets 1$ \KwTo{} $M$}{
            $w^i_j \gets w^{i-1}$\tcp*[l]{Copy global to client}
            $a_j \gets 0$\;
            \For{$k \gets 1$ \KwTo{} $n$}{
                \tcp{Train one minibatch on client}
                $\mathcal M \gets$ sample minibatch\;
                $\mathcal L \gets \text{ComputeLoss}(\mathcal M; w^i_j)$
                $w^i_j \gets w^i_j - \eta \nabla \mathcal L$\;
                $a_j \gets a_j + |M|$\;
            }
        }
        $w^i \gets \frac{1}{\sum_{l=1}^M a_l} \sum_{j=1}^M  a_j w^i_j$\tcp*[l]{Update global}
    }
\end{algorithm}

\subsubsection{FL for Condition Monitoring and Fault Diagnosis}
An immediately apparent benefit of FL for manufacturing
is the ability to train on multiple data sets
without exposing sensitive factory data to the server.
Motivated by this privacy perspective,
\cite{zhang2021_kbs} proposed FL for building
a fault diagnosis model from isolated data sets,
although the method assumes all clients see matching faults.
Client models with low validation performance are ignored
when aggregating the global model to improve robustness.
A peer-to-peer adaptation of FL showed improvements over
local training at each node
for detecting wind turbine and bearing faults~\cite{wang2021_mssp}.
\cite{xia2022} also investigated FL for bearing fault diagnosis
while proposing a vertical FL algorithm based on gradient tree boosting
to accommodate clients with different feature subsets.
For Remaining Useful Life (RUL) applications,
\cite{du2023} implemented FL for collaborative training of
transformer models on degradation data from simulated turbofan aircraft engines.

\subsubsection{Multi-Party and Single-Party Incentives for FL}
Beyond privacy, FL offers benefits to both coalitions of multiple manufacturers
and within a single, distributed manufacturer.
In additive manufacturing,
\cite{mehta2022_jms}
found that FL improves defect image segmentation over locally trained client models
and showed that performance gains can both
incentivize manufacturers to join existing federations
and incentivize these federations to welcome new clients.
Work by \cite{deng2022_jim}
further supports FL's ability to
improve model performance versus locally trained models
while preserving privacy among aircraft manufacturers.
Even if manufacturers decline federations with competitors
to avoid possible model poisoning~\cite{ding2023_tie},
FL offers substantial benefits for communication-efficient
training on distributed data owned by a single manufacturing entity,
reducing the network traffic needed
to maximize utilization of distributed sensing.
However, in both the multi-party and single-party paradigms,
FL implementations must handle discrepancies between clients
while still maximally leveraging a collaborative approach.

\subsubsection{FL for Heterogeneous Clients}
In practical applications,
clients could have different tasks or distributions of data,
making basic \texttt{FedAvg} suboptimal for each member
but still desirable for privacy benefits.
Initializing FL clients with a pretrained global feature extractor
can reduce the required training time on individualized downstream tasks~\cite{wang2021_tii}.
However, the case studies only tested this for image domain tasks.
Similarly, a personalized FL approach
can locally optimize feature extractors
and classifiers while penalizing shifts between the local classifier weights
and the globally optimized weights~\cite{shi2022}.
This permits the clients to share information without a hard constraint that fixes weights among them.
Surprisingly, if the clients observe different faults,
\cite{mehta2023b} demonstrated that FL can share classifier information
across rotating machinery clients even if they have unbalanced or non-i.i.d classes.
Injecting noise and creating fake pseudoclasses within each client
can also help with globally aligning classes between models~\cite{li2023}.
Conversely, if the client input distributions differ significantly,
a single global model might not be successful.
\cite{mehta2023a} opted to cluster gradient updates from members
and perform FL separately within each subgroup.
Experiments validated the algorithm on benchmark data and a custom bearing fault data set.
However, these studies in heterogeneous FL critically stop short
of addressing the problem of massively unlabeled data at each client.
Furthermore, when the number of observed classes is extremely limited,
relying supervised learning could hinder the discriminativeness of learn representations.


\section{Proposed Methods for Maximizing Model Generalization\label{sec:70-proposed}}

\begin{figure*}[t]
    \centering
    \includegraphics[width=\textwidth]{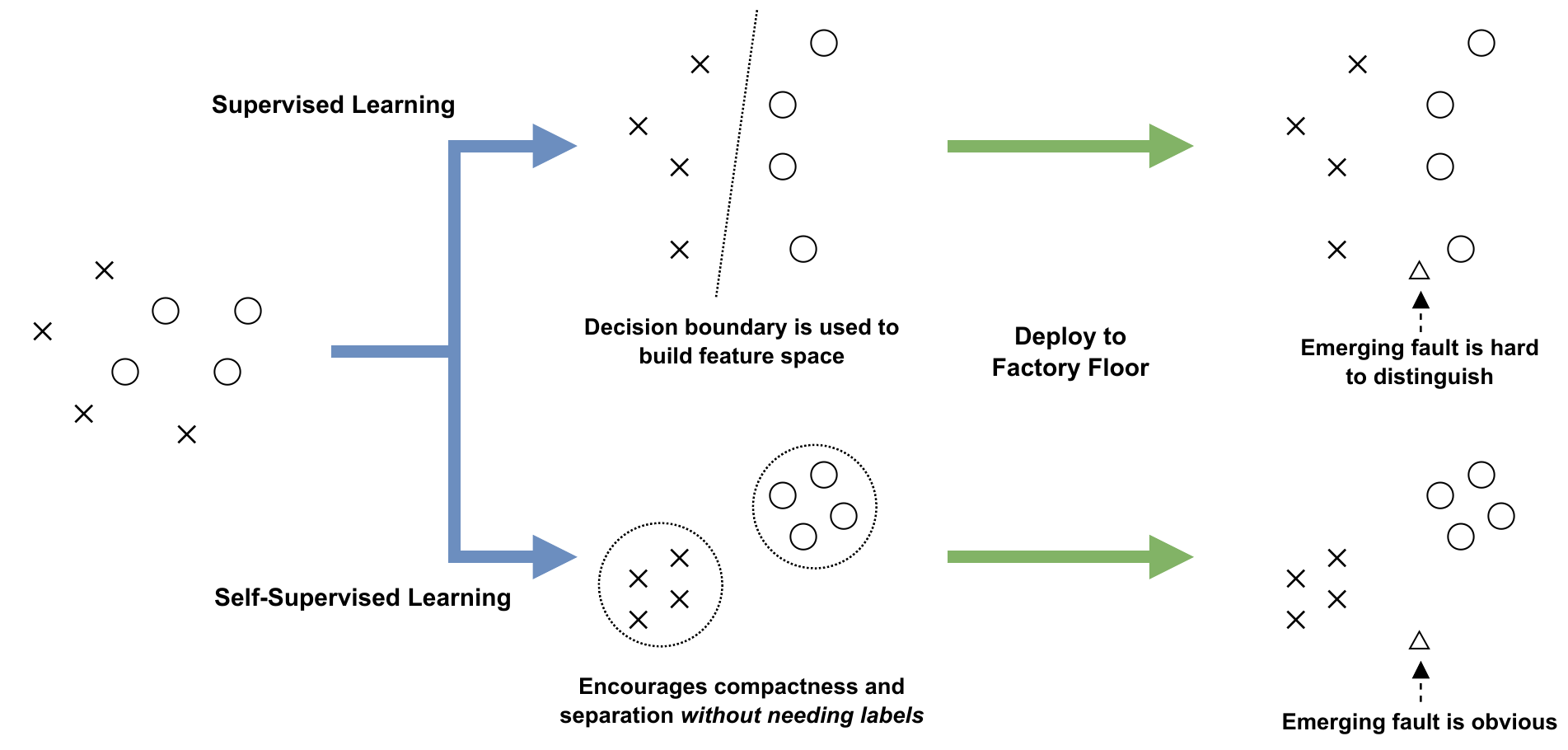}
    \caption{SSL encourages compactness and separation of pseudoclasses
        while supervised representations are dependent on decision boundaries.\label{fig:70-sl-vs-ssl}}
\end{figure*}

\begin{figure*}[t]
    \centering
    \includegraphics[width=\linewidth]{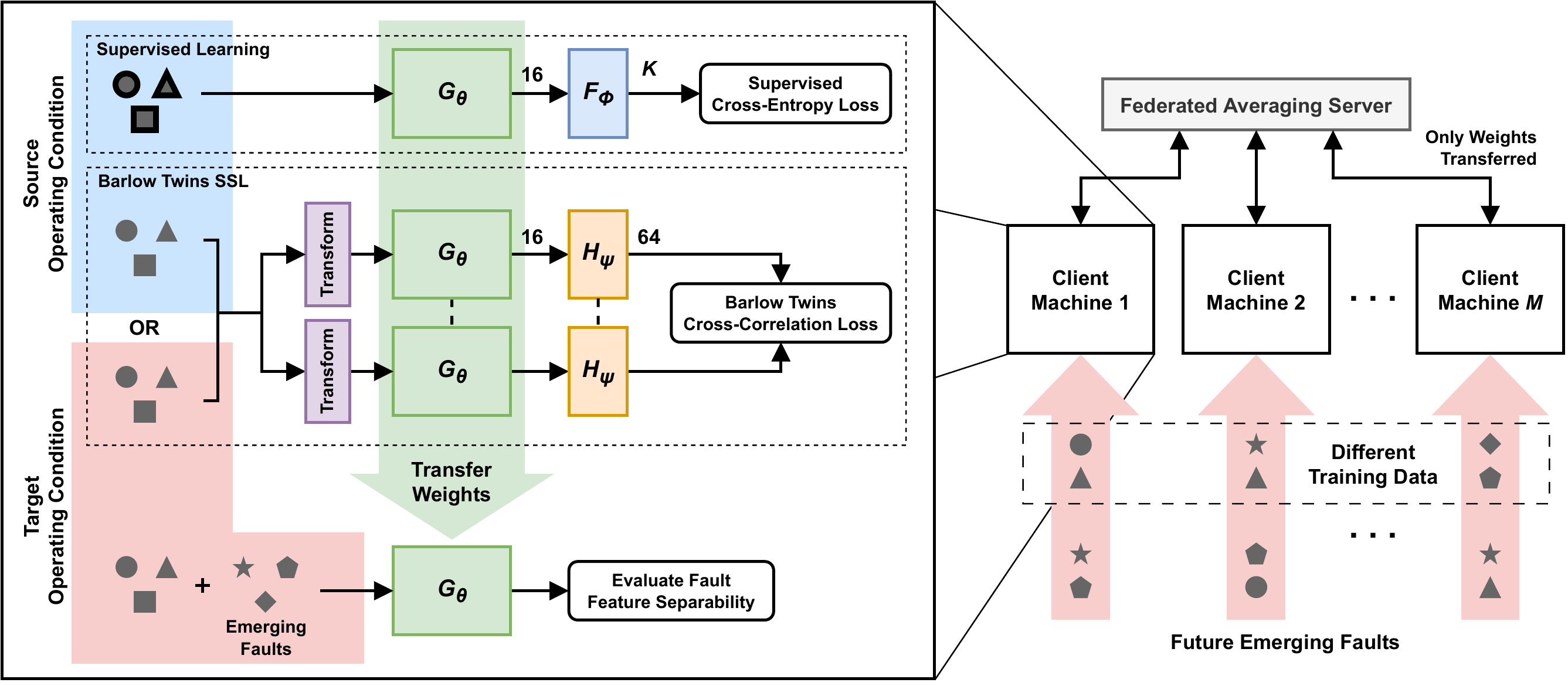}
    \caption{Proposed methods for comparing
        the discriminability of emerging faults
        when transferring weights from a supervised
        or self-supervised 1D CNN feature extraction backbone.
        Federated Learning can then be used to share information
        efficiently among multiple client machines.\label{fig:70-overview}}
\end{figure*}

Although supervised learning on massively diverse data sets
may produce generalizable features~\cite{tian2020_eccv},
it may struggle
when class (i.e., fault/condition) diversity is limited in two ways by
1) producing less compact clusters,
and 2) allowing noise or systematic biases to dominant feature extraction.
A simple classification objective
constructs the feature space and 
decision boundaries
without explicitly encouraging compact clusters
(see Fig.~\ref{fig:70-sl-vs-ssl}).
With limited training classes,
the model has few decision boundaries
with which to partition the feature space.
This could produce loosely structured features,
increasing the likelihood that features
from future emerging faults will overlap
those from previous health conditions.
While adding compactness objectives may help,
fewer classes also means the model has fewer observations
from varied environmental conditions and process parameters.
Since DL implementations are free to learn features themselves,
a supervised model could resort to
systematic biases to separate data
rather than the more complex underlying fault signals as intended.
Combining
data from distributed machines
could mitigate these issues by increasing class diversity,
but
aggregating
high-velocity sensing streams
could be
difficult given
bandwidth constraints.
Furthermore, most raw data will be unlabeled regardless,
making large-scale supervised learning impossible.
The proposed method
instead
adopts SSL to
support unlabeled data and
improve the feature space structure
and FL to expand the effective data set size without
inundating communication networks
or introducing privacy concerns
(see Fig.~\ref{fig:70-overview}).
Together, these techniques
learn a more discriminative feature space that generalizes
to new operating conditions and emerging faults.

\subsection{Barlow Twins}

\begin{algorithm}[t]
    \caption{Random augmentations for Barlow Twins
        in PyTorch style\label{alg:70-augmentations}}
    \footnotesize
    \begin{Verbatim}[commandchars=\\\{\}]
    \PY{c+c1}{\PYZsh{} x: 1D input tensor with shape (B, C, L)}
    \PY{k}{def} \PY{n+nf}{randomly\PYZus{}augment}\PY{p}{(}\PY{n}{x}\PY{p}{)}\PY{p}{:}
        \PY{c+c1}{\PYZsh{} Random jitter}
        \PY{n}{jitter} \PY{o}{=} \PY{n}{random}\PY{o}{.}\PY{n}{randrange}\PY{p}{(}\PY{n}{x}\PY{o}{.}\PY{n}{shape}\PY{p}{[}\PY{o}{\PYZhy{}}\PY{l+m+mi}{1}\PY{p}{]}\PY{p}{)}
        \PY{n}{x} \PY{o}{=} \PY{n}{torch}\PY{o}{.}\PY{n}{cat}\PY{p}{(}
            \PY{p}{(}\PY{n}{x}\PY{p}{[}\PY{p}{:}\PY{p}{,} \PY{p}{:}\PY{p}{,} \PY{n}{jitter}\PY{p}{:}\PY{p}{]}\PY{p}{,} \PY{n}{x}\PY{p}{[}\PY{p}{:}\PY{p}{,} \PY{p}{:}\PY{p}{,} \PY{p}{:}\PY{n}{jitter}\PY{p}{]}\PY{p}{)}\PY{p}{,}
            \PY{n}{dim}\PY{o}{=}\PY{o}{\PYZhy{}}\PY{l+m+mi}{1}\PY{p}{,}
        \PY{p}{)}
        \PY{c+c1}{\PYZsh{} Random scale}
        \PY{n}{vmax} \PY{o}{=} \PY{p}{(}
            \PY{n}{x}\PY{o}{.}\PY{n}{abs}\PY{p}{(}\PY{p}{)}
            \PY{o}{.}\PY{n}{reshape}\PY{p}{(}\PY{n}{x}\PY{o}{.}\PY{n}{shape}\PY{p}{[}\PY{l+m+mi}{0}\PY{p}{]}\PY{p}{,} \PY{o}{\PYZhy{}}\PY{l+m+mi}{1}\PY{p}{)}
            \PY{o}{.}\PY{n}{max}\PY{p}{(}\PY{n}{dim}\PY{o}{=}\PY{o}{\PYZhy{}}\PY{l+m+mi}{1}\PY{p}{,} \PY{n}{keepdim}\PY{o}{=}\PY{k+kc}{True}\PY{p}{)}\PY{p}{[}\PY{l+m+mi}{0}\PY{p}{]}
        \PY{p}{)}
        \PY{n}{max\PYZus{}scale} \PY{o}{=} \PY{n}{vmax}\PY{o}{.}\PY{n}{reciprocal}\PY{p}{(}\PY{p}{)}
        \PY{n}{min\PYZus{}scale} \PY{o}{=} \PY{l+m+mf}{0.1}
        \PY{n}{scales} \PY{o}{=} \PY{p}{(}
            \PY{n}{torch}\PY{o}{.}\PY{n}{rand\PYZus{}like}\PY{p}{(}\PY{n}{max\PYZus{}scale}\PY{p}{)}
            \PY{o}{*} \PY{p}{(}\PY{n}{max\PYZus{}scale} \PY{o}{\PYZhy{}} \PY{n}{min\PYZus{}scale}\PY{p}{)}
            \PY{o}{+} \PY{n}{min\PYZus{}scale}
        \PY{p}{)}
        \PY{n}{x} \PY{o}{=} \PY{n}{x} \PY{o}{*} \PY{n}{scales}\PY{o}{.}\PY{n}{unsqueeze}\PY{p}{(}\PY{o}{\PYZhy{}}\PY{l+m+mi}{1}\PY{p}{)}
        \PY{c+c1}{\PYZsh{} Random mask}
        \PY{n}{mask\PYZus{}size} \PY{o}{=} \PY{l+m+mi}{64}
        \PY{n}{mask\PYZus{}start} \PY{o}{=} \PY{n}{random}\PY{o}{.}\PY{n}{randrange}\PY{p}{(}
            \PY{n}{x}\PY{o}{.}\PY{n}{shape}\PY{p}{[}\PY{o}{\PYZhy{}}\PY{l+m+mi}{1}\PY{p}{]} \PY{o}{\PYZhy{}} \PY{n}{mask\PYZus{}size}
        \PY{p}{)}
        \PY{n}{mask\PYZus{}end} \PY{o}{=} \PY{n}{mask\PYZus{}start} \PY{o}{+} \PY{n}{mask\PYZus{}size}
        \PY{n}{x}\PY{p}{[}\PY{p}{:}\PY{p}{,} \PY{p}{:}\PY{p}{,} \PY{n}{mask\PYZus{}start}\PY{p}{:}\PY{n}{mask\PYZus{}end}\PY{p}{]} \PY{o}{=} \PY{l+m+mf}{0.0}
    
        \PY{k}{return} \PY{n}{x}
    \end{Verbatim}
\end{algorithm}

\begin{figure}[t]
    \centering
    \includegraphics[width=2.5in]{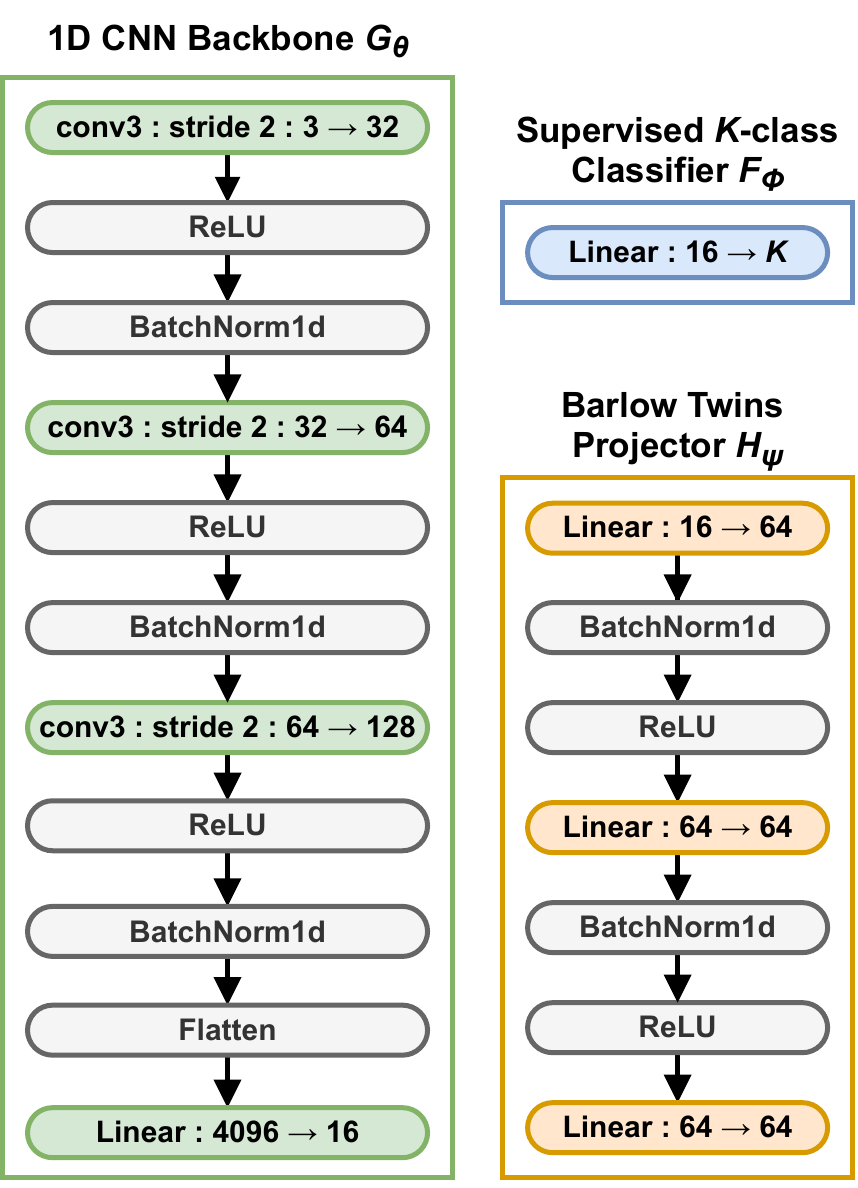}
    \caption{The architectures for the
        1D CNN backbone feature extractor $G_\theta$,
        supervised $K$-class classifier $F_\phi$,a
        and Barlow Twins projection head $H_\psi$.\label{fig:70-models}}
\end{figure}

Replacing supervised learning with SSL
introduces the knowledge-informed assumption that
although emerging faults or new operating conditions
have not been observed,
this time series data from the target domain
will have similar building blocks
and salient characteristics---e.g., frequency content---that
discriminates them.
To extract these salient indicators instead of unwanted biases,
SSL relies on expert-designed random data augmentations
that indicate the expected variation with the signals.
Barlow Twins SSL seeks to tightly cluster feature projections
from different augmentations of the same observation
by maximizing the cross-correlation between projections.
This ensures that examples falling within the expected signal variation
are grouped closely together.
The augmentations
themselves should be informed by knowledge of condition monitoring signals
to randomize unimportant signal attributes
while preserving the semantic class~\cite{peng2022_tie}.
Extending the proposed augmentations from \cite{russell2023_tie},
Algorithm~\ref{alg:70-augmentations} outlines the random transformations
used with Barlow Twins
in the proposed methods for condition monitoring.
The examples are randomly shifted (jittered) in time,
scaled, and masked.
Given input batch $X$ of $n$ examples,
feature extraction backbone $G_\theta$,
projector $H_\psi$,
Barlow Twins first computes the projections of two augmented versions
$X'$ and $X''$ of the input batch (according to Algorithm~\ref{alg:70-augmentations})
and their corresponding projections
$Z' = H_\psi(G_\theta(X'))$ and $Z'' = H_\psi(G_\theta(X''))$.
Then both sets of projections are normalized across the batch:
\begin{equation}
    \label{eq:70-barlow-twins-normalization}
    \begin{aligned}
        \mu_i       & = \frac 1 n \sum_{k=1}^n Z_{ik}               \\
        \sigma_i^2  & = \frac 1 n \sum_{k=1}^n {(Z_{ik} - \mu_i)}^2 \\
        \hat Z_{ij} & = (Z_{ij} - \mu_i) / \sigma_i                 \\
    \end{aligned}
\end{equation}
Next, the cross-correlation matrix $R$ is computed and normalized by the batch size:
\begin{equation}
    \label{eq:70-barlow-twins-cross-correlation}
    R = \hat{Z}' \hat{Z}''^\top / n
\end{equation}
Finally, the loss function can be calculated using $R$:
\begin{equation}
    \label{eq:70-barlow-twins-loss}
    \mathcal L_{\text{BT}}(R) =
    \mathrm{tr}\left({(R - I)}^2\right) + \lambda \sum_i \sum_{j \neq i} R_{ji}
\end{equation}
where $\lambda$ controls the strength of the independence constraint.
The first term encourages the diagonal elements to be one,
meaning that
individual
features are highly correlated (aligned) across the batch,
meaning that instances within the expected variation---as defined by
the applied random augmentations--will
map to similar feature projections (i.e., cluster together).
The second term drives off-diagonal elements to zero so
each feature is independent from the rest.
This improves the representational capacity
by ensuring multiple features do not encode the same information.
With this loss function,
the Barlow Twins feature extractor and projection head
can be trained with standard stochastic gradient descent
and backpropation methods.
Fig.~\ref{fig:70-models} shows the architecture of the
1D CNN backbone $G_\theta$ for extracting features
from condition monitoring data
and the Barlow Twins projection head $H_\psi$.

\subsection{Federated Learning for Information Sharing}

\begin{figure*}[t]
    \centering
    \includegraphics[width=\textwidth]{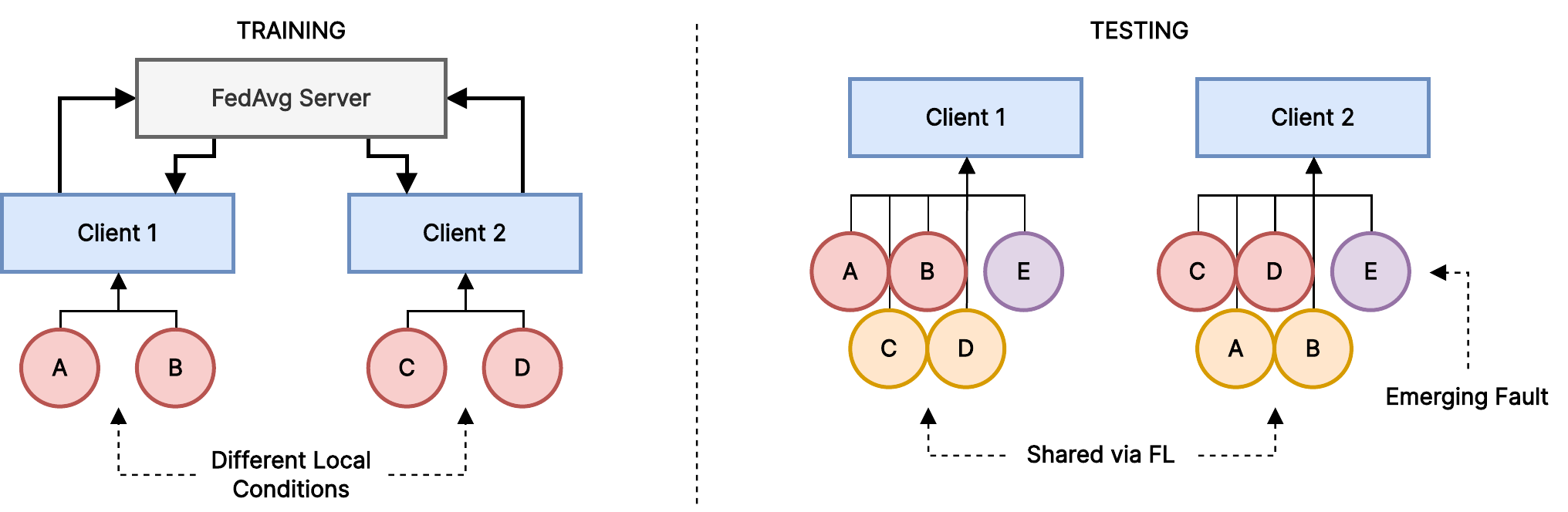}
    \caption{Each client experiences different conditions,
        and averaging model weights diffuses this knowledge to other clients,
        maximizing the diversity of the data set and improving performance on emerging faults.\label{fig:info-share}}
\end{figure*}

Most factory floors will have multiple similar machines
that will each experience different health conditions throughout operation.
Data from a single machine may contain very few distinct conditions,
but network contraints may prevent each machine from streaming
all its sensing data to the cloud to construct a unified data set.
The machines themselves may not be geographically colocated
or may belong to separate manufacturers without data-sharing agreements.
To circumvent these hindrances,
the model can be trained with \texttt{FedAvg}
(see Algorithm~\ref{alg:70-fedavg}).
Each client machine retains complete ownership of its data
while indirectly gaining knowledge about new health conditions
through model averaging
on the FL server.
This indirect information sharing between clients via the global model
can be viewed as a form of TL.
When each client receives an updated global model,
they benefit from the observations and knowledge of the other clients.
Thus, even if a client lacks training experience with a given health condition,
if another client \textit{has} trained with that condition,
the FL algorithm will diffuse this experience back to the uninformed client
(see Fig.~\ref{fig:info-share}).
Thus, FL may offer TL advantages among the clients,
improving the generalization of each one to future fault conditions.
Moreover, The client machines only send updated models
to the FL server once per round,
significantly reducing the volume and velocity of data
transmitted to the cloud.
By combining FL with SSL,
DL can operate in realistic condition monitoring scenarios
with unlabeled, distributed training data
while reducing network communication and maintaining manufacturer privacy.

\section{Experiments\label{sec:70-experiments}}

Two case studies investigate the proposed claims.
The first compares the generalizability of representations
after pretraining with supervised learning or SSL
on varying numbers of distinct classes.
The second examines the impact of distributed training with FL
on model performance under emerging faults.

\subsection{Motor Condition Data Set}

\begin{figure*}[t]
    \centering
    \includegraphics[width=\linewidth]{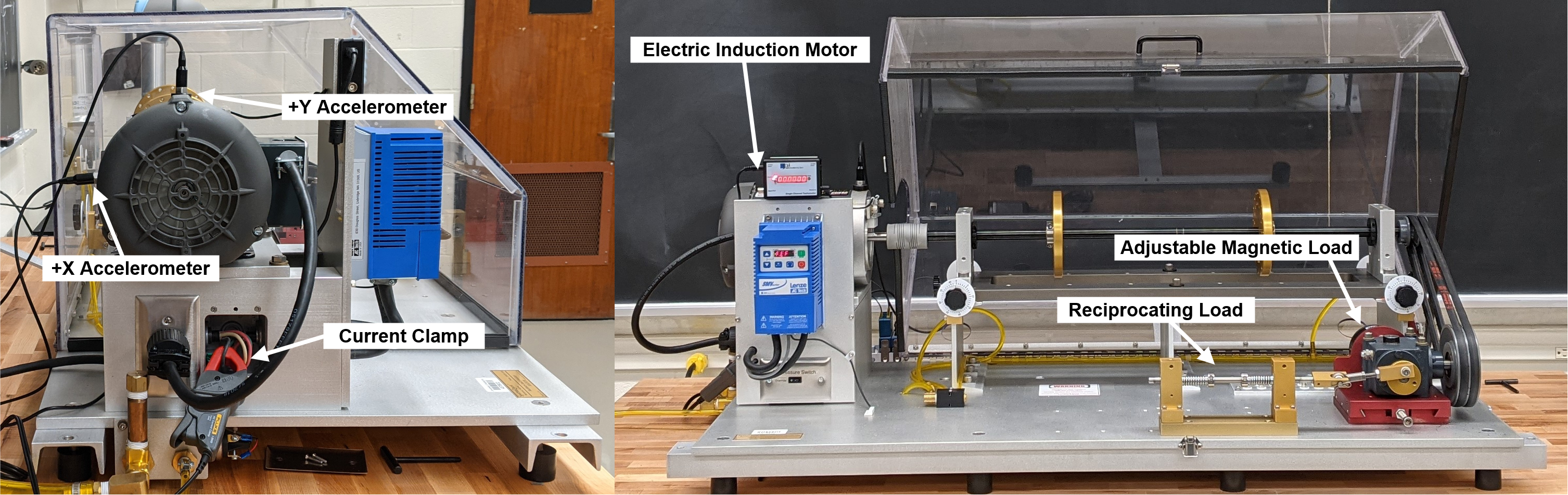}
    \caption{The SpectraQuest Machinery Fault Simulator
        used to collect the motor health condition
        data set.\label{fig:70-mfs}}
\end{figure*}

Both case studies use a motor fault condition data set
collected from the SpectraQuest Machinery Fault Simulator (MFS)
in Fig.~\ref{fig:70-mfs}.
With a 12 kHz sampling rate,
two accelerometers mounted orthogonally capture vibration data,
and a current clamp measures electrical current signals.
Sixty seconds of steady-state data is gathered
for eight motor health conditions:
normal (N), faulted bearings (FB),
bowed rotor (BoR), broken rotor (BrR),
misaligned rotor (MR), unbalanced rotor (UR),
phase loss (PL), and unbalanced voltage (UV).
Each of the conditions is run at 2000 RPM and 3000 RPM
with loads of 0.06 N$\cdot$m and 0.7 N$\cdot$m
for a total of 32 unique combinations
of health conditions and process parameters.
For simplicity, each unique combination can be identified
with \textit{xy}
where \textit{x} is 2 or 3
to specify the RPM parameter,
and \textit{y} is ``H'' or ``L'' to specify a high or low load parameter
(e.g., 3L refers to 3000 RPM with load of 0.06 N$\cdot$m).
The signals are then normalized to $[-1, 1]$
and split into 256-point windows for the DL experiments.

\subsection{Transfer Learning Experiments}

The first set of experiments tests the claim
that SSL is a more effective TL pretraining method.
The experimental design reflects the following assumptions:
\begin{enumerate}
    \itemsep0em
    \item labeled training data is available
          from a source set of process parameters,
    \item unlabeled training data is available
          from a target set of process parameters, and
    \item the pretrained model may encounter new fault types once deployed.
\end{enumerate}
This scenario leads to
three comparison methods:
\begin{itemize}
    \item \textbf{Supervised (Source)}:
          supervised training on the labeled source domain data
    \item \textbf{Barlow Twins (Source)}:
          self-supervised training on the source domain data (ignoring labels)
    \item \textbf{Barlow Twins (Target)}:
          self-supervised training on the unlabeled target domain data
\end{itemize}
All three methods use the same 1D CNN feature extraction backbone $G$
shown in Fig.~\ref{fig:70-models}.
The supervised network adds the $K$-class classifier $F_\phi$ to the backbone,
while Barlow Twins adds the projection head $H_\psi$.
The networks $F_\phi$ and $G_\theta$ are then optimized
using stochastic gradient descent
and backpropagation with cross-entropy loss
from (\ref{eq:70-cross-entropy-loss}).
The Barlow Twins model produces projections
$Z' = H_\psi(G_\theta(X'))$ and $Z'' = H_\psi(G_\theta(X''))$
from input batch augmentations $X'$ and $X''$
(see Algorithm~\ref{alg:70-augmentations}),
and the training loss is computed from
(\ref{eq:70-barlow-twins-normalization})--(\ref{eq:70-barlow-twins-loss})
with $\lambda = 0.01$.
Both the supervised and self-supervised models are trained for 1000 epochs
with an Adam optimizer and learning rate of 0.0005.

To assess the quality and generalizability of each method's representation,
the frozen features of each pretrained network
are used to train a privileged linear evaluation classifier
with access to labeled target domain data from all eight health conditions
(the \textit{evaluation data set}),
following conventions in the literature
for evaluating SSL models~\cite{zbontar2021_icml}.
Access to privileged label information prevents
this classifier from being trained and deployed in practice,
but it follows the accepted standard for assessing the separability
of the underlying feature representations.
The evaluation classifier is trained for 75 epochs on the frozen features,
and the test set accuracy is used to judge the representation quality.

To simulate the occurrence of new, unseen faults,
the source and target domain training data sets
are limited to two, four, or six randomly selected health conditions.
Since the evaluation data set contains all eight conditions,
this corresponds to encountering six, four, or two
previously unseen classes after pretraining,
respectively.

\begin{table}[t]
    \centering
    \caption{Transfer Learning Health Condition Sets\label{tab:70-tl-cond-sets}}
    \begin{tabular}{cr}
        \toprule
        \# of Conditions & Condition Classes                       \\
        \midrule
        2                & \makecell[r]{\{N, PL\}                  \\
        \{PL, BoR\}                                                \\
        \{BrR, UV\}                                                \\
        \{UR, UV\}                                                 \\
        \{FB, UV\}}                                                \\
        \midrule
        4                & \makecell[r]{\{N, BrR, UR, UV\}         \\
        \{PL, BrR, MR, UV\}                                        \\
        \{FB, PL, BoR, UV\}                                        \\
        \{FB, BrR, MR, UR\}                                        \\
        \{BoR, BrR, MR, UR\}}                                      \\
        \midrule
        6                & \makecell[r]{\{N, FB, PL, BrR, MR, UR\} \\
        \{N, PL, BoR, MR, UR, UV\}                                 \\
        \{N, PL, BoR, BrR, MR, UV\}                                \\
        \{N, FB, BoR, BrR, MR, UV\}                                \\
        \{N, PL, BoR, BrR, MR, UR\}}                               \\
        \bottomrule
    \end{tabular}
\end{table}

To capture variation caused by the source/target domain selection,
training health conditions,
and model initialization,
450 experiments are conducted,
150 for each of the three comparative methods.
The 150 runs come from all combinations of two source/target domain pairs
(3L$\to$2H or 2H$\to$3L),
15 unique health condition configurations for the source/target training data,
and five random seeds (0 through 4).
The 15 combinations of training health conditions consist of
five randomly sampled sets for each of two, four, and six health conditions
(see Table~\ref{tab:70-tl-cond-sets}).
All experiments use an NVIDIA V100 GPU with 32 GB of RAM
for hardware acceleration.

\subsection{Federated Learning Experiments}

\begin{table}[t]
    \centering
    \caption{Federated Learning Health Condition Sets\label{tab:70-fl-cond-sets}}
    \begin{tabular}{crr}
        \toprule
        ID & Client 1    & Client 2    \\
        \midrule
        1  & \{BoR, MR\} & \{BrR, UR\} \\
        2  & \{FB, UR\}  & \{BrR, UV\} \\
        3  & \{BoR, N\}  & \{BrR, FB\} \\
        4  & \{BrR, UV\} & \{UR, N\}   \\
        5  & \{FB, MR\}  & \{BoR, UV\} \\
        \bottomrule
    \end{tabular}
\end{table}

The FL experiments determine whether sharing model information
between clients with disjoint sets of training conditions
will improve the distinguishability of future emerging faults.
To evaluate this,
two clients are each assigned two randomly selected motor health conditions.
Each client has local training data for its two conditions
from all process parameters combinations (i.e., 2L, 2H, 3L and 3H).
The FL server provides both clients
with an initial global model with random weights.
In each round of FL, the clients train their local model
on their unique set of two health conditions
and then return the updated model to the server.
The server averages the weights
and redistributes the new model to the clients
in preparation for the next round of FL (see Algorithm~\ref{alg:70-fedavg}).

FL experiments are run for 1000 rounds,
and each client trains for 20 local batches
in each round.
When performing supervised learning, each client updates the weights
using cross-entropy loss from (\ref{eq:70-cross-entropy-loss}).
For Barlow Twins training,
each client uses the cross-correlation loss from
(\ref{eq:70-barlow-twins-normalization})--(\ref{eq:70-barlow-twins-loss}).
Both supervised learning and Barlow Twins use
the same network architectures for TL shown in Fig.~\ref{fig:70-overview}
and are trained with an Adam optimizer and learning rate of 0.0002.

Each of the four possible model
configurations---supervised learning and Barlow Twins
each with and without FL---is trained
with five random seeds (0 through 4) to gauge variation
caused by random initialization.
Five unique sets of training conditions are tested to marginalize effects
of individual health conditions (see Table~\ref{tab:70-fl-cond-sets}).
All combinations of the four methods, five seeds, and five condition sets
lead to a total of 100 FL experiments.
All experiments use an NVIDIA V100 GPU for hardware acceleration.
Similar to TL, both clients are evaluated
using the accuracy of a privileged linear classifier
trained on the frozen feature extraction network
to classify all eight conditions.
The classifier is trained for 75 epochs after FL is complete.

\section{Results and Discussion\label{sec:70-results}}
The results indicate that Barlow Twins
produces more generalizable and transferable representations
than supervised learning,
and that FL for information sharing may further improve performance.

\subsection{Transfer Learning Results}

\begin{figure}[t]
    \centering
    \includegraphics[width=3.25in]{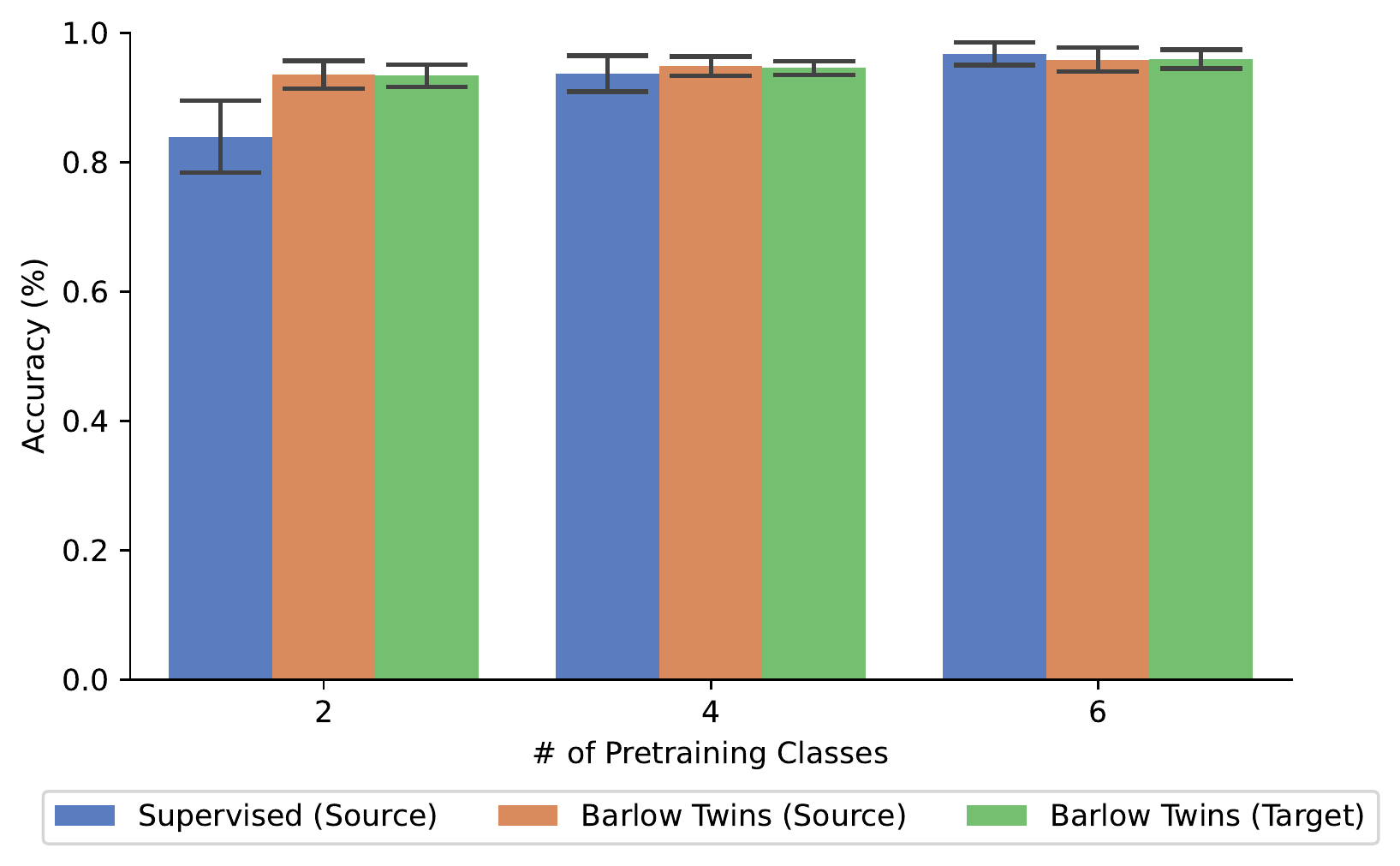}
    \caption{Target domain accuracy of the weight transfer methods
        on all eight motors condition
        versus number of faults
        in the training domain.\label{fig:70-tl-bar-graph}}
\end{figure}

\begin{figure*}[t]
    \centering
    \begin{subfigure}[b]{0.47\textwidth}
        \centering
        \includegraphics[width=\textwidth]{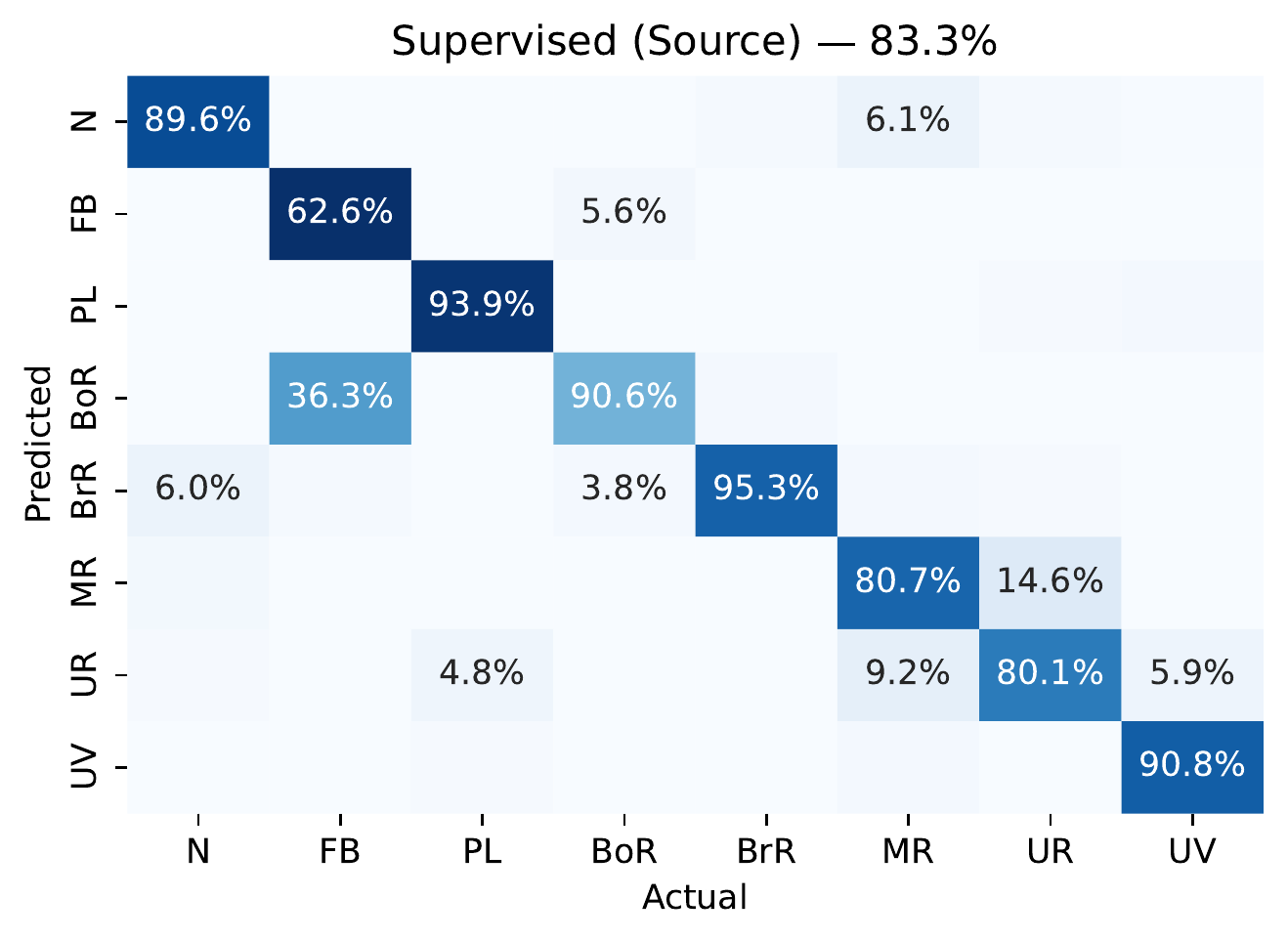}
        \caption{Supervised (Source)}
    \end{subfigure}
    \hfill
    \begin{subfigure}[b]{0.47\textwidth}
        \centering
        \includegraphics[width=\textwidth]{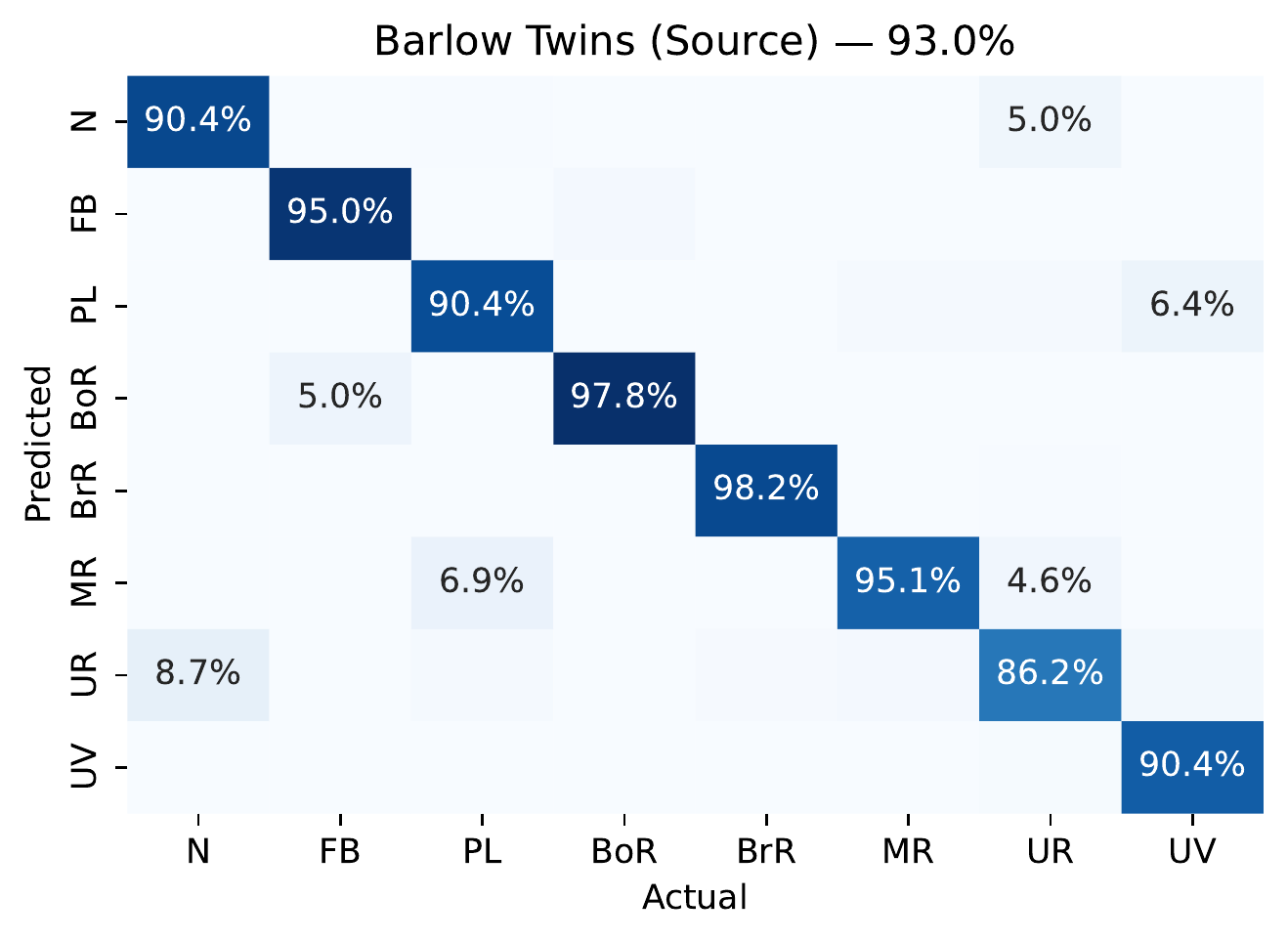}
        \caption{Barlow Twins (Source)}
    \end{subfigure}
    \caption{Representative confusion matrices
        showing the advantage of using Barlow Twins
        over supervised learning
        when transferring models to new process parameters (3L$\to$2H)
        with six emerging conditions\label{fig:tl-cm}}
\end{figure*}

Table~\ref{tab:70-tl-results} and Fig.~\ref{fig:70-tl-bar-graph} present the key TL results
comparing supervised learning on labeled source process parameters,
Barlow Twins on unlabeled source process parameters,
and Barlow Twins on unlabeled target process parameters.
The accuracy metrics are computed from the test split of the evaluation data set
containing all eight conditions under the target process parameters.
Even when just two conditions are available for training,
Barlow Twins generates a separable representation
capable of 93.5\% accuracy when shown all eight health conditions.
In the same scenario, supervised learning is limited to 83.9\% accuracy.
Fig.~\ref{fig:tl-cm} shows representative confusion matrices
that highlight the improvements of SSL over supervised learning.
For example, supervised learning struggles to distinguish
the misaligned rotor (MR) and unbalanced rotor (UR) conditions
while using Barlow Twins boosts the accuracy within these categories
by 15 and 6 points, respectively.
The SSL approach still confuses some classes
(e.g., N$\leftrightarrow$UR and UR$\to$\{MR,N\})
possibly because the random augmentations could not
fully span the expected variation of these classes.
That is, data labeled as normal varied more
than the expected variation captured by the random jitter, scaling, and masking from Algorithm~\ref{alg:70-augmentations}.
As a result, Barlow Twins clustered some of these examples closer to UR,
leading the evaluation classifier to miscategorize them.
Similarly, some PL examples are classified as MR using Barlow Twins representations,
indicating that these PL instances experience some variation that clustered them closer to MR.
Future work can investigate whether these examples truly are miscategorized
or actually resemble members of the confused class.
Barlow Twins can also utilize unlabeled target domain data
to further improve the representation---Barlow Twins (Target)
in Table~\ref{tab:70-tl-results}---while
supervised learning cannot use this data due to the lack of labels.
Interestingly, Barlow Twins (Target)
does not show a clear improvement over Barlow Twins (Source)
indicating that SSL
is effective for learning generalizable features from the motor condition monitoring source domain data.

As more conditions are included in training,
the performance convergence of supervised learning and Barlow Twins
can be explained according to the optimization objective of each approach.
Supervised learning seeks to split the data
along decision boundaries for the classifier.
While this may ensure the training classes are distinguishable,
it does not guarantee compactness of the feature clusters.
Thus, it is suspected that features from new, emerging faults could overlap
with those from faults seen in training.
In contrast, Barlow Twins encourages similar input instances
to have correlated and closely matching features.
This emphasis on feature similarity produces tight clusters
that reduce the likelihood of new fault features
overlapping with existing clusters.
When the number of training conditions increases,
the additional decision boundaries created by supervised learning
naturally improve feature cluster compactness,
bringing its evaluation accuracy closer to that of Barlow Twins.
However, because manufacturing applications will have limited class diversity
compared to the possible number of emerging faults,
these results show the general superiority of SSL-based representations
over those transfered from supervised learning in uncertain operating environments.

\begin{table}[t]
    \centering
    \caption{Transfer Learning Evaluation Accuracy Results (\%)\label{tab:70-tl-results}}
    \begin{tabular}{lrrr}
        \toprule
                                      & \multicolumn{3}{c}{\# of Training Health Conditions}                                                                   \\
        Method                        & 2                                                    & 4                              & 6                              \\ 
        \midrule
        Supervised {\small(Source)}   & 83.9$\pm$5.6                                         & 93.7$\pm$2.8                   & \textbf{96.7}$\pm$\textbf{1.8} \\ 
        Barlow Twins {\small(Source)} & \textbf{93.5}$\pm$\textbf{2.1}                       & \textbf{94.8}$\pm$\textbf{1.5} & 95.8$\pm$1.9                   \\ 
        Barlow Twins {\small(Target)} & 93.3$\pm$1.7                                         & 94.5$\pm$1.1                   & 95.9$\pm$1.4                   \\ 
        \bottomrule
    \end{tabular}
\end{table}

\subsection{Federated Learning Results}

\begin{figure}[t]
    \centering
    \includegraphics[width=3.25in]{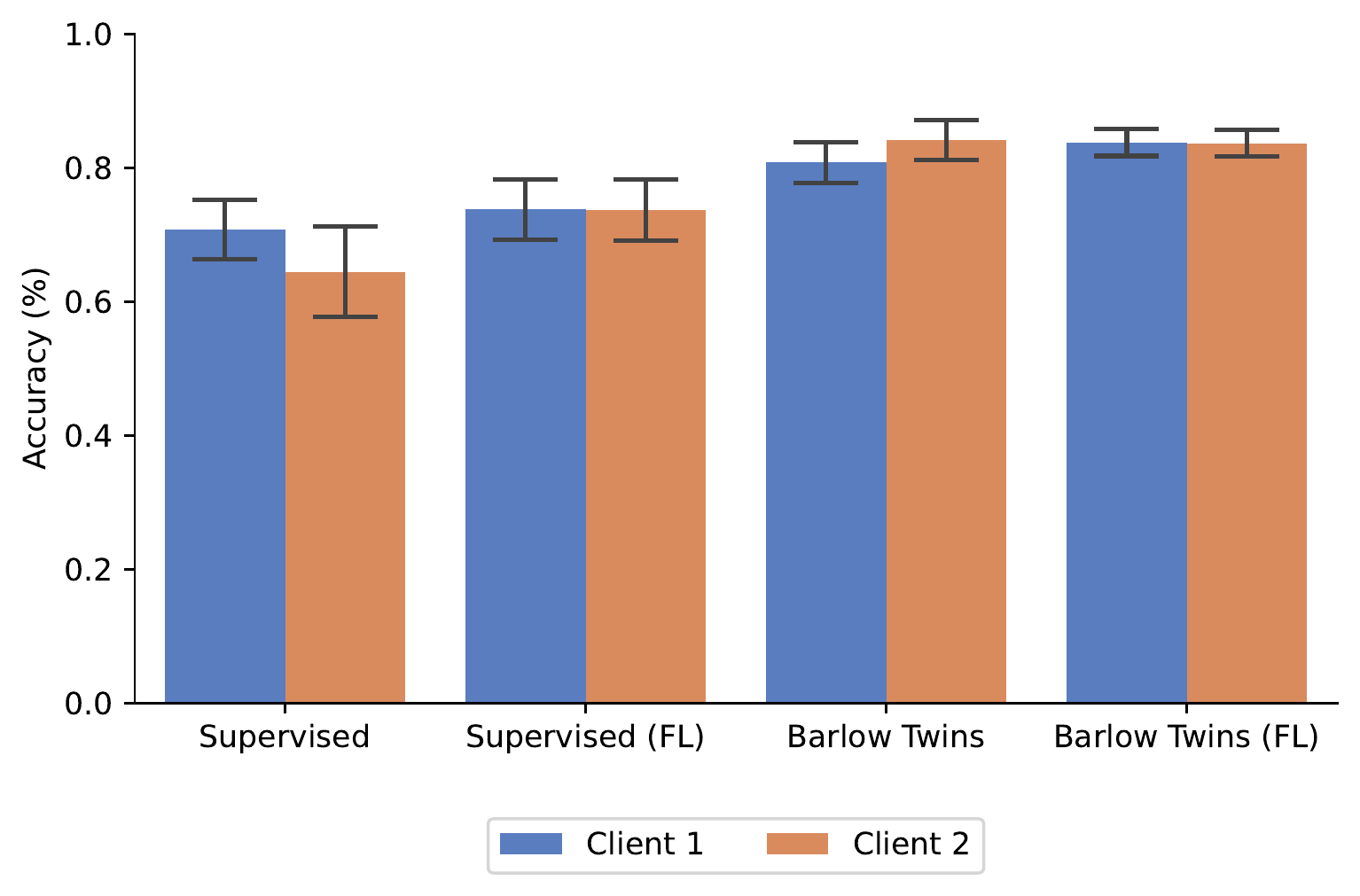}
    \caption{Client evaluation accuracies
        on all health conditions\label{fig:70-fl-bar-graph}}
\end{figure}

Table~\ref{tab:70-fl-results} and Fig.~\ref{fig:70-fl-bar-graph}
present the FL results.
Supervised learning shows an noticeable increase
in discriminability of emerging faults when FL is included.
Without FL, the overall evaluation accuracy between the clients is only 67.6\%.
When FL is included, information about the health conditions
is shared indirectly through the \texttt{FedAvg} server,
boosting the overall accuracy to 73.7\%.
Since both clients share a global model during FL,
they have nearly identical accuracy.
When trained without FL,
the supervised learning clients show a 6-point discrepancy.

\begin{figure*}[t]
    \centering
    \begin{subfigure}[b]{0.47\textwidth}
        \centering
        \includegraphics[width=\textwidth]{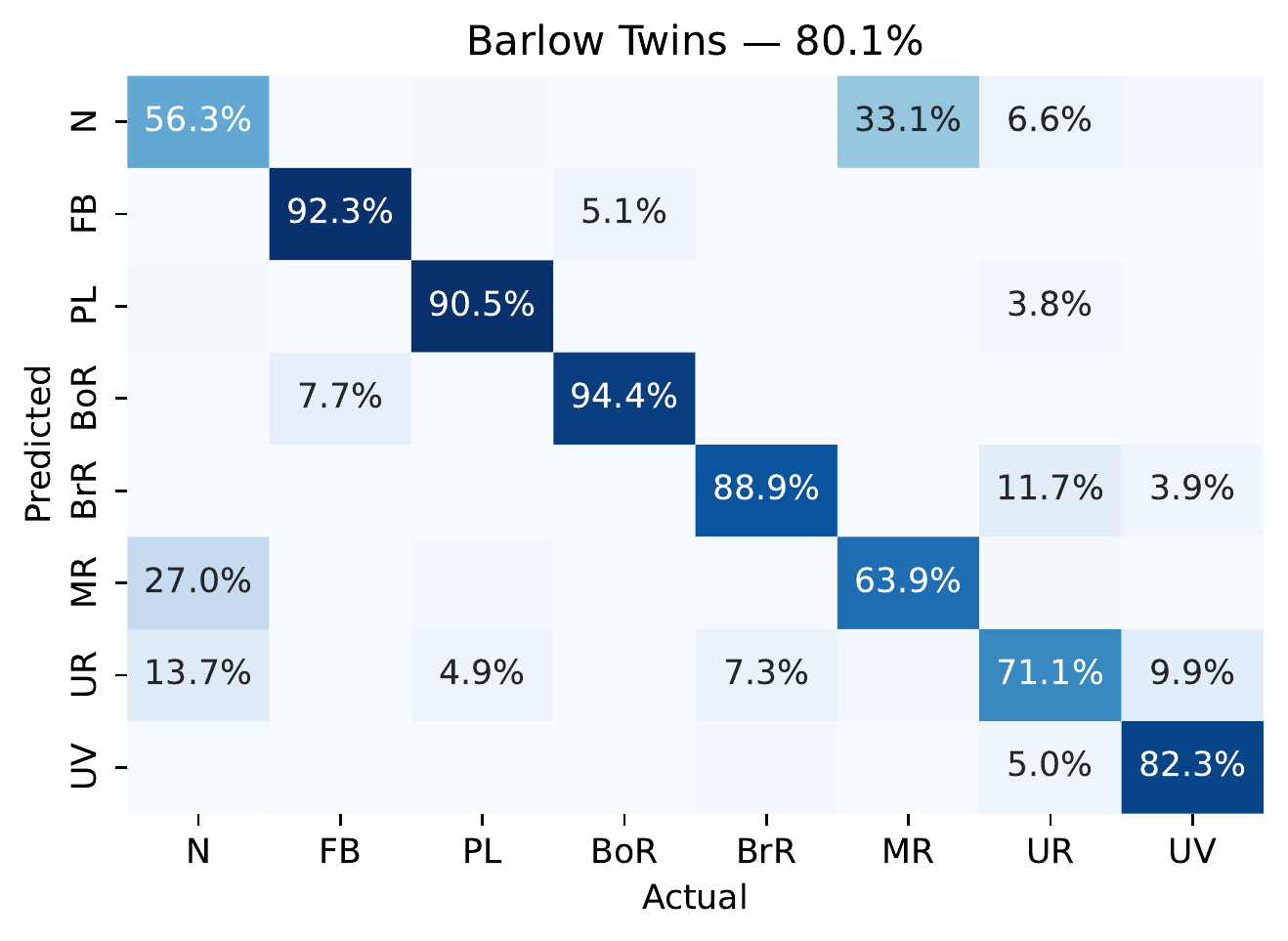}
        \caption{Without FL}
    \end{subfigure}
    \hfill
    \begin{subfigure}[b]{0.47\textwidth}
        \centering
        \includegraphics[width=\textwidth]{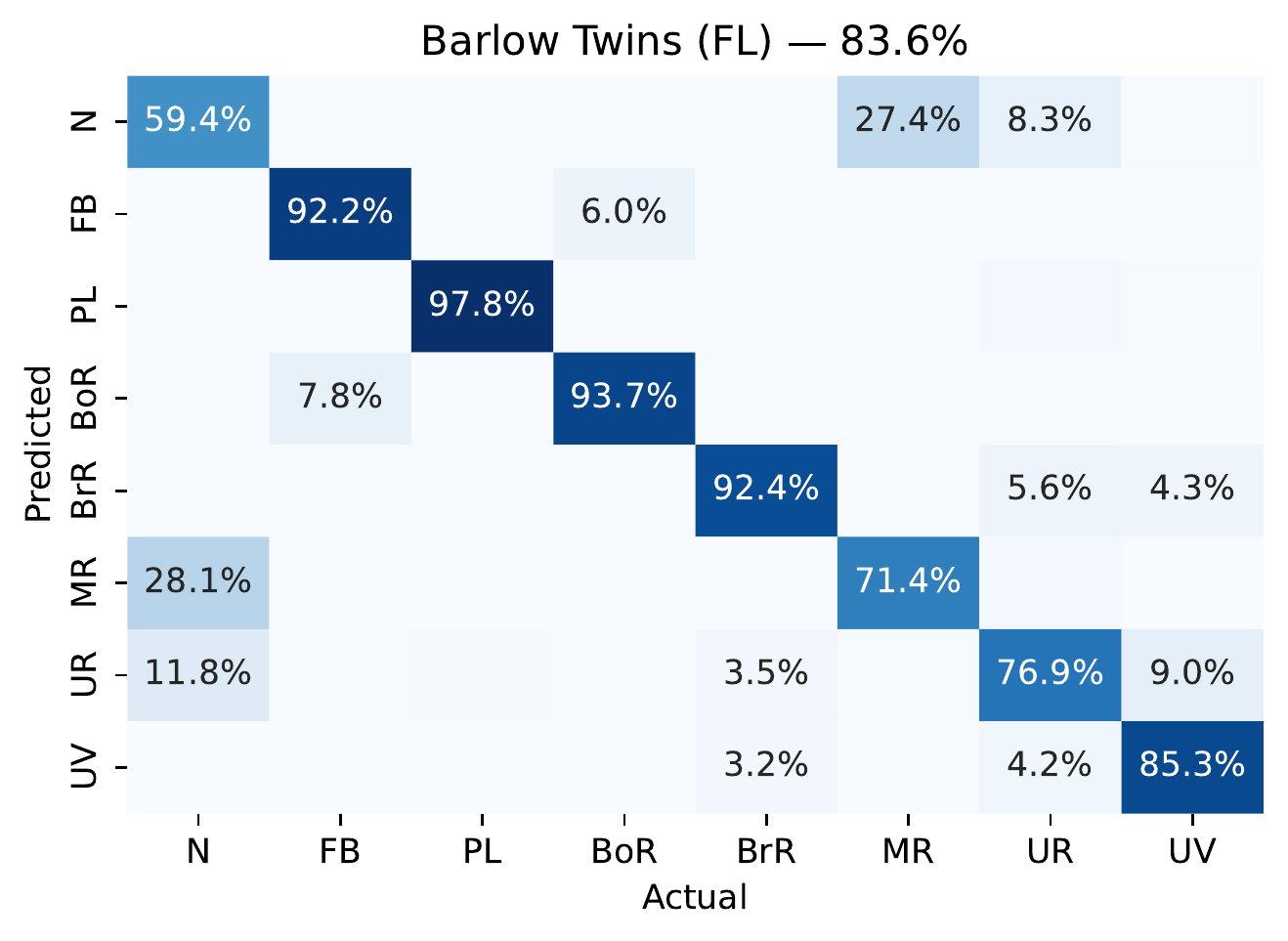}
        \caption{With FL}
    \end{subfigure}
    \caption{Representative confusion matrices
        showing the benefits of including FL for Barlow Twins Client 1.
        Client 1 was trained on \{BoR, N\},
        while Client 2 (not shown) was trained on \{BrR, FB\}\label{fig:fl-ssl-cm}}
\end{figure*}

Barlow Twins outperforms all
supervised learning methods even when FL is excluded.
The separately-trained clients reach an overall evaluation accuracy of 82.4\%.
Once FL combined with Barlow Twins,
performance increases to 83.7\%,
the highest overall accuracy among all methods.
As in the supervised case,
FL also reduces the discrepancy between the clients,
reducing the accuracy difference from 3.3 points to 0.1 point.
The representative confusion matrices in Fig.~\ref{fig:fl-ssl-cm}
show in the improvement in Client 1 when FL is included.
Phase loss (PL) accuracy increases from 90.5\% to 97.8\%,
and misaligned rotor (MR) accuracy increases from 63.9\% to 71.4\%.
The differences in accuracy with respect to the TL-only results
may be a result of training with all sets of process parameters
instead of a single source domain set.
The Barlow Twins data augmentations might be effective for one or two process parameter sets,
they require additional development to capture the class variation expected
across all the process parameter combinations.
For example, the TL experiments might more easily distinguish N vs. MR and UR
because the transfer occurred between 2H$\leftrightarrow$3L
naturally leading to more distant clusters
than when data contains only a single process parameter change.
While these are directions for future work,
these
preliminary
results demonstrate how indirect information sharing
through the \texttt{FedAvg} server
may be able to boost discriminability of emerging faults,
if the individual clients see a limited number
of distinct health conditions.
By merging models trained on different subsets of health conditions,
FL may increase the diversity of the training data set,
improving the generalization of the learned features.

\begin{table}[t]
    \centering
    \caption{Federated Learning Accuracy Results (\%)\label{tab:70-fl-results}}
    \begin{tabular}{lrrr}
        \toprule
        Method            & Client 1                       & Client 2                       & Overall                        \\
        \midrule
        Supervised        & 70.7$\pm$4.5                   & 64.4$\pm$6.8                   & 67.6$\pm$6.5                   \\
        Supervised (FL)   & 73.8$\pm$4.5                   & 73.7$\pm$4.6                   & 73.7$\pm$4.5                   \\
        \midrule
        Barlow Twins      & 80.8$\pm$3.1                   & \textbf{84.1}$\pm$\textbf{3.0} & 82.4$\pm$3.5                   \\
        Barlow Twins (FL) & \textbf{83.7}$\pm$\textbf{2.0} & 83.6$\pm$2.0                   & \textbf{83.7}$\pm$\textbf{2.0} \\
        \bottomrule
    \end{tabular}
\end{table}

\subsection{Limitations and Future Directions}
While the results are promising,
the case study focuses on motor condition monitoring.
Additional experiments are necessary to validate
the approach more fully against a variety of manufacturing problems and data sets.
In addition, while SSL facilitates learning discriminative representations without labeled data,
downstream classification tasks still require an additional step
to either cluster features automatically into presumed class groups
or integrate a human-in-the-loop solution in which an operator can tag
a limited number of features with labels.
Future work should also characterize when SSL and FL approaches
struggle with manufacturing data.
Understanding possible shortcomings and failure modes
will enable practitioners to rapidly implement the right technology for a given problem.

\section{Conclusion\label{sec:70-conclusion}}
Given growing developments in SSL,
this study compares the generalization
of feature representations learned via SSL
versus those learned via supervised methods.
In weight transfer experiments,
a feature extractor trained with Barlow Twins
outperformed a supervised classifier
when transferring to an operating environment
with different process parameters
that contained emerging faults.
With only two health conditions for training,
the features learned by Barlow Twins
from the source domain
produced an evaluation classifier accuracy
9.6 points higher than that of
the representation learned
by supervised training on labeled source domain data.
To further improve peformance,
knowledge of distributed but similar
SSL client models
can inform an FL architecture
that shares fault experience
while respecting privacy concerns.
Thus, manufacturing applications
with large unlabeled data sets
can use SSL and FL to learn generalizable representations
for emerging faults even without diverse, labeled data.
With enhanced emerging fault detection across conditions,
models will be better equipped for the factory floor
and improve the trustworthiness
and reliability
of practical condition monitoring deployments.

\section*{Acknowledgements}
This work is supported by the National Science Foundation under Grant No. 2015889.
We would thank the University of Kentucky Center for Computational Sciences and
Information Technology Services Research Computing for their support and use of
the Lipscomb Compute Cluster and associated research computing resources.

\bibliographystyle{elsarticle-num-names}
\bibliography{ref}
\end{document}